\documentclass{SIP}
\usepackage{graphicx}
\usepackage{amsmath}
\usepackage{multirow}
\usepackage{threeparttable}
\usepackage{amsfonts}
\usepackage{mathrsfs}
\usepackage{arydshln}
\usepackage{enumitem}
\usepackage{url}
\usepackage{afterpage}
\usepackage{booktabs}
\usepackage{tabularx}
\usepackage{algorithm}
\usepackage{algorithmicx}
\usepackage{algpseudocode}
\usepackage{bm}
\usepackage{comment}

\makeatletter

\let\MYcaption\@makecaption
\makeatother
\usepackage{subcaption}
\makeatletter
\let\@makecaption\MYcaption
\makeatother
\newcommand{\myvector}[1]{\boldsymbol{#1}}
\newcommand{\mymatrix}[1]{\mathrm{#1}}

\newcommand{\ip}[2]{\langle #1, #2 \rangle}

\newcounter{num}

\begin{document}

\title{Self-Supervised Intrinsic Image Decomposition Network Considering Reflectance Consistency}

\author{Yuma Kinoshita and Hitoshi Kiya}

\address{Tokyo Metropolitan University, Tokyo, Japan}

\corres{\name{Hitoshi Kiya}
\email{kiya@tmu.ac.jp}}

\begin{abstract}
We propose a novel intrinsic image decomposition network
considering reflectance consistency.
Intrinsic image decomposition aims to decompose an image
into illumination-invariant and illumination-variant components,
referred to as ``reflectance'' and ``shading,'' respectively.
Although there are three consistencies that the reflectance
and shading should satisfy,
most conventional work does not sufficiently account
for consistency with respect to reflectance,
owing to the use of a white-illuminant decomposition model
and the lack of training images capturing
the same objects under various illumination-brightness and -color conditions.
For this reason,
the three consistencies are considered in the proposed network
by using a color-illuminant model
and training the network with losses calculated
from images taken under various illumination conditions.
In addition,
the proposed network can be trained in a self-supervised manner
because various illumination conditions can easily be simulated.
Experimental results show that our network can decompose images
into reflectance and shading components.
\end{abstract}

\keywords{Authors should not add keywords, as these will be chosen during the submission process .}

\maketitle

\section{Introduction}
  Decomposing a natural image into illumination-invariant and -variant components
  enables us to easily modify the image.
  For example, since a low-light image can be considered as
  an image having lower illumination-variant components
  than those of the corresponding well-exposed image,
  it can be enhanced by 
  amplifying its illumination-variant components~\cite{fu2016weighted,guo2017lime,chien2019retinex}.
  Similarly,
  white balancing can be done by
  changing the color of illumination-variant components.
  For this reason, intrinsic image decomposition~\cite{barrow1978recovering, grosse2009groundtruth}
  has so far been studied to decompose a natural image into two such components.

  Intrinsic image decomposition is based on the Retinex theory~\cite{land1977retinex}.
  In this theory,
  a natural image consists of the \textit{reflectance} and \textit{shading} of a scene,
  where the reflectance and the shading correspond to the illumination-invariant component
  and the illumination-variant component, respectively.
  To enable the decomposition, various methods have so far been proposed
  ~\cite{land1977retinex, weiss2001deriving, bell2014intrinsic, zhou2015learning, ma2018single, fan2018revisiting,
  li2018cginstrinsics, li2018learning, lettry2018unsupervised,
  wang2019single, liu2020unsupervised}.
  An early attempt at intrinsic image decomposition used image-gradient information~\cite{land1977retinex}.
  The gradient-based method estimates reflectance by thresholding image gradients
  under the assumption that large gradients are due to spatial variations in reflectance
  and small gradients are due to light shading.
  However, the effectiveness of the gradient-based method is limited because it is heuristic.
  For this reason, optimization-based decomposition methods that utilize sparse modeling or Bayesian inference
  have also been widely studied
  ~\cite{shen2011intrinsic, chang2014bayesian, bell2014intrinsic}.
  In addition, recent methods for intrinsic image decomposition are based on
  deep neural networks (DNNs).
  These DNN-based methods significantly improve the performance of the decomposition
  compared with conventional gradient- or optimization-based approaches,
  by supervised learning
  ~\cite{fan2018revisiting, zhou2015learning, li2018learning, wang2019single}.

  In intrinsic image decomposition,
  there are three premises regarding consistency:
  reconstruction consistency,
  reflectance consistency in terms of illumination brightness,
  and reflectance consistency in terms of illumination colors.
  Most conventional methods for intrinsic image decomposition use a white-illuminant model
  that does not consider illumination color,
  and thus consider only a part of these premises,
  i.e., reconstruction consistency and reflectance consistency (brightness).
  In such a case,
  reflectance components decomposed by these methods are affected
  by illumination-color conditions.
  Some conventional methods~\cite{li2018learning} consider all of the premises
  by using a color-illuminant model that considers illumination color.
  In these methods, DNNs are trained by using a highly-synthetic dataset or a human-labeled dataset of a real scene
  ~\cite{grosse2009groundtruth, butler2012naturalistic, bell2014intrinsic}.
  However, such datasets are insufficient to generalize real scenes.
  Although unsupervised decomposition methods have recently been proposed
  ~\cite{chien2019retinex, liu2020unsupervised, lettry2018unsupervised,
  ma2018single, li2018cginstrinsics},
  they still have limited performance.

  To solve these problems, in this paper,
  we propose a novel intrinsic image decomposition network
  that considers both all three premises and the problem with data.
  To consider the premises,
  we use a color-illuminant model and
  train the network with losses calculated by using image sets,
  where each image set consists of images capturing
  the same objects under various illuminant-brightness and -color conditions.
  In addition, we will show that such an image set can easily be generated from a single raw image.
  Namely, our network can be trained in a self-supervised manner
  by using a general raw image dataset.
  Therefore, the difficulty in preparing a large amount of data can be overcome.
  Since the proposed network can perform intrinsic image decomposition
  so that the resulting reflectance is unaffected by the illumination conditions,
  the proposed network will be useful for image enhancement while preserving object color
  and more realistic white balance adjustment.

  We evaluate the performance of the proposed intrinsic image decomposition network
  in terms of the robustness against illumination-brightness/-color changes.
  To measure the robustness,
  we utilize images taken under various illumination-brightness and -color conditions
  and decompose them into reflectance and shading components.
  By comparing the reflectance components for the images
  on the basis of the peak signal-to-noise ratio (PSNR),
  mean squared error (MSE),
  and structural dissimilarity (DSSIM),
  the robustness against illumination-brightness/-color changes can be evaluated.
  Experimental results show that the proposed decomposition network is
  robust against illumination-brightness/-color changes.
  In addition,
  we will discuss the effectiveness of the proposed network through a simulation experiment
  using the MIT intrinsic images dataset~\cite{grosse2009groundtruth}.

\section{Preliminaries}
\label{sec:preliminaries}
  In this section, we briefly summarize intrinsic image decomposition and its issues.
  We use the notations shown in Table \ref{tab:notation} throughout this paper.
  \begin{table}[t]
    \centering
    \footnotesize
    \caption{Notation \label{tab:notation}}
    \begin{tabular}{lc}\hline\hline
      Symbol & Definition \\ \hline
      $a$ & A scalar \\
      $\myvector{a}$ & A vector \\
      $\mymatrix{A}$ & A matrix \\
      $H, W$ & Image height and weight, respectively \\
      $x, y$ & A pixel coordinate \\
      $P$ & Set of pixels of an image, i.e., $\{(x, y)\}$ \\
      $\myvector{I}(x, y)$ & An RGB vector at pixel $(x, y)$ of an image \\
      $\myvector{R}(x, y)$ & An RGB vector at pixel $(x, y)$ of a reflectance \\
      $S(x, y)$ & A scalar value at pixel $(x, y)$ of a shading in gray scale \\
      $\myvector{c}$ & An RGB vector indicating illumination color \\
      $\myvector{S}(x, y)$ & An RGB vector at pixel $(x, y)$ of a shading \\
     \hline
    \end{tabular}
  \end{table}

\subsection{Intrinsic image decomposition \label{subsec:models}}
\begin{figure}[t]
    \centering
    \includegraphics[width=\columnwidth]{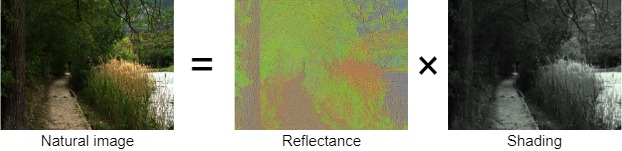}
    \caption{Intrinsic image decomposition.
      In accordance with Retinex theory,
      natural image can be written as pixel-wise product of
      illumination-invariant component (i.e., reflectance)
      and illumination-variant component (i.e., shading).
      \label{fig:retinex}}
  \end{figure}

  The goal of intrinsic image decomposition is to decompose a given image
  into a pixel-wise product of an illuminant-dependent component called \textit{reflectance}
  and an illuminant-independent component called \textit{shading} as shown in Fig. \ref{fig:retinex}.

  Here, there are two main models for intrinsic image decomposition as follows.
  \begin{description}[leftmargin=*]
    \item [White-illuminant model],
      ~which describes the relationship among an RGB image,
      reflectance, and shading as
      \begin{equation}
        \myvector{I}(x, y) = S(x, y) \myvector{R}(x, y),
        \label{eq:white_model}
      \end{equation}
      where $(x, y)$ indicates a pixel coordinate.
      $\myvector{I}(x, y)$ and $\myvector{R}(x, y)$ are
      3-dimensional RGB vectors for the image and reflectance, respectively,
      and $S(x, y)$ is a scalar value for shading.
      This model assumes the illumination color is white because $S(x, y)$ is a scalar value.
    \item [Color-illuminant model],
      ~which describes the relationship among an RGB image,
      reflectance, and shading as
      \begin{equation}
        \myvector{I}(x, y) = \myvector{S}(x, y) \odot \myvector{R}(x, y),
        \label{eq:color_model}
      \end{equation}
      where $\myvector{I}(x, y), \myvector{S}(x, y)$, and $\myvector{R}(x, y)$ are
      3-dimensional RGB vectors for the image, reflectance, and shading, respectively.
      This model considers illumination color because $\myvector{S}(x, y)$ is an RGB vector.
  \end{description}

\subsection{Scenario \label{subsec:scenario}}
  In intrinsic image decomposition, there are three premises:
  \begin{description}[style=nextline]
    \item[Reconstruction consistency]
      The product of the estimated reflectance and shading matches the corresponding original image,
      as shown in Eq. (\ref{eq:white_model}) or Eq. (\ref{eq:color_model}).
    \item[Reflectance consistency (brightness)]
      Reflectances are invariant against a change in illumination brightness.
    \item[Reflectance consistency (color)]
      Reflectances are invariant against a change in illumination colors.
  \end{description}

  Most conventional work on intrinsic image decomposition uses the white-illuminant model
  in Section \ref{sec:preliminaries}.\ref{subsec:models})
  and thus considers only a part of these premises,
  i.e., reconstruction consistency and reflectance consistency (brightness).
  However, the white-illuminant model cannot satisfy
  reflectance consistency (color)
  because illumination color is not considered in this model.
  In such a case,
  reflectance components decomposed by these methods are affected
  by illumination-color conditions.

  For this reason,
  the color-illuminant model is used in recent research work.
  In \cite{li2018learning}, all three premises are considered by training a DNN
  by using videos taken by a camera at a fixed location.
  However, the DNN still has limited performance
  due to there being a limited amount of real data for training.

  Because of such situations, in this paper,
  we propose a novel deep intrinsic image decomposition network that considers
  all three premises and the problem with data.
  To consider the premises,
  losses are calculated by using image sets,
  where each image set consists of raw images capturing
  the same objects under various illumination-brightness and -color conditions.
  In addition, we will show that such an image set can easily be generated from a single raw image.
  Namely, our network can be trained in a self-supervised manner
  by using a general raw image dataset.
  Therefore, the problem with the amount of data can be overcome.

\section{Proposed intrinsic image decomposition network}
  In this paper,
  we aim to decompose an image into reflectance and shading
  by using a deep neural network.
  The key idea of our approach is to utilize
  image sets where each set consists of images capturing
  the same objects under various illuminant-brightness and -color conditions
  in consideration of the three premises
  in Section \ref{sec:preliminaries}.\ref{subsec:scenario}).
  We will show that such an image set can easily be generated from a single raw image,
  in Section \ref{sec:self_supervised}.

\subsection{Overview}
  For the proposed method, we consider the color-illumination model in Eq. (\ref{eq:color_model}).
  Additionally, we assume that the illumination color of an image is the same for all pixels.
  As a result, shading is written as
  \begin{align}
    \myvector{S}(x, y) &= S(x, y) \myvector{c},
    \label{eq:our_model}
  \end{align}
  where $S(x, y)$ is shading in gray scale,
  and $\myvector{c}$ is an RGB vector indicating the illumination color.
  Hence, the goal of the proposed method is to obtain
  reflectance $\myvector{R}$, gray-shading $S$, and RGB vector $\myvector{c}$
  from given input image $\myvector{I}$.

\begin{figure}[t]
    \centering
    \includegraphics[width=\columnwidth]{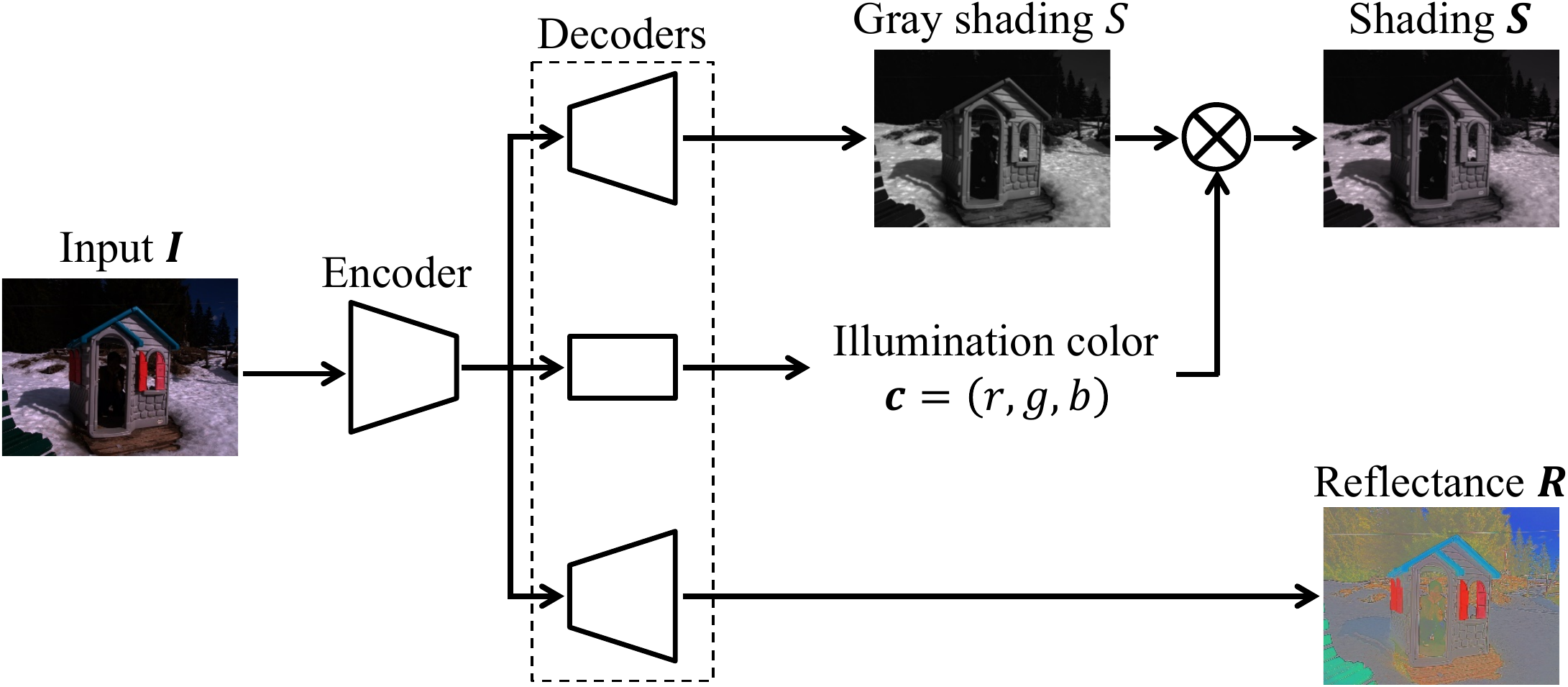}
    \caption{Proposed intrinsic image decomposition. Encoder maps given input image
    into features. Features are fed into three decoders: reflectance decoder,
    shading decoder, and illumination-color decoder.
    Reflectance/shading decoders produce reflectance/gray-shading components, respectively.
    Illumination-color decoder estimates illumination color.
    Colored shading can be obtained by multiplying gray-shading and illumination-color.
    \label{fig:overview}}
  \end{figure}

  Figure \ref{fig:overview} illustrates the architecture of the proposed network.
  Our network has a single encoder and three decoders.
  With the encoder, input image $\myvector{I}$ is transformed into feature maps
  that are fed into decoders.
  An estimation $\hat{\myvector{R}}$ of reflectance in RGB color space
  is directly obtained as an output of a reflectance decoder.
  In contrast, an estimation $\hat{\myvector{S}}$ of shading is given as
  the product of estimated illumination color $\hat{\myvector{c}}$ and estimated shading $\hat{S}$ in gray scale.
  $\hat{\myvector{c}}$ and $\hat{S}$ are obtained by an illumination-color decoder and a shading decoder, respectively.

\subsection{Loss functions}
  Let the $n$-th image set in the training dataset be $\{ \myvector{I} \}_n$
  To fulfill all three premises in Section \ref{sec:preliminaries}.\ref{subsec:scenario}),
  our network is trained to minimize loss $\mathcal{L}$ calculated by using 
  $\myvector{I}_1, \myvector{I}_2 \in \{ \myvector{I} \}_n$ as
  \begin{equation}
    \mathcal{L}
      = \mathbb{E}\left[
            \mathcal{L}_{\mathrm{reconst}}(\myvector{I}_1, \myvector{I}_2)
          + \mathcal{L}_{\mathrm{reflect}}(\myvector{I}_1, \myvector{I}_2)
        \right],
    \label{eq:all_loss}
  \end{equation}
  where $\mathcal{L}_{\mathrm{reconst}}$ and $\mathcal{L}_{\mathrm{reflect}}$ are loss functions
  for considering reconstruction and reflectance consistencies, respectively.

  For the reconstruction consistency,
  in accordance with Eq. (\ref{eq:color_model}), 
  we use $\mathcal{L}_{\mathrm{reconst}}$ given as
  \begin{align}
    \mathcal{L}_{\mathrm{reconst}}(\myvector{I}_1, \myvector{I}_2)
      &= \lambda_1 \left( \mathcal{L}_{\ell_1}(\myvector{I}_1, \hat{\myvector{I}}_1)
        + \mathcal{L}_{\ell_1}(\myvector{I}_2, \hat{\myvector{I}}_2) \right) \nonumber \\
      &- \lambda_2 \left( \mathcal{L}_{\cos}(\myvector{I}_1, \hat{\myvector{I}}_1)
        + \mathcal{L}_{\cos}(\myvector{I}_2, \hat{\myvector{I}}_2) \right),
    \label{eq:recon_loss}
  \end{align}
  where $\hat{\myvector{I}}_i, i \in \{1, 2\}$ is an image reconstructed
  from estimated reflectance $\hat{\myvector{R}}_i$,
  gray-scale shading $\hat{S}_i$,
  and illumination color $\hat{\myvector{c}}_i$ for image $I_i$.
  From Eqs. (\ref{eq:color_model}) and (\ref{eq:our_model}),
  $\hat{\myvector{I}}_i$ is calculated as
  \begin{equation}
    \hat{\myvector{I}}_i(x, y) = \left( \hat{S}(x, y)\hat{\myvector{c}}_i \right) \odot \hat{\myvector{R}}_i(x, y).
    \label{eq:reconstruction}
  \end{equation}
  $\mathcal{L}_{\ell_1}$ is the mean absolute error,
  and $\mathcal{L}_{\cos}$ is cosine similarity given by
  \begin{equation}
    \mathcal{L}_{\cos}(\myvector{I}, \hat{\myvector{I}})
      = \frac{1}{|P|} \sum_{(x, y) \in P}
        \frac{\ip{\myvector{I}(x, y)}{\hat{\myvector{I}}(x, y)}}{\|\myvector{I}(x, y)\|_2 \cdot \|\hat{\myvector{I}}(x, y)\|_2}.
  \end{equation}

  For the reflectance consistencies,
  we designed $\mathcal{L}_{\mathrm{reflect}}$ as 
  \begin{align}
    \mathcal{L}_{\mathrm{reflect}}(\myvector{I}_1, \myvector{I}_2)
      &= \lambda_3 \mathcal{L}_{\ell_1}(\hat{\myvector{R}}_1, \hat{\myvector{R}}_2) \nonumber \\
      &+ \lambda_4 \left( \mathcal{L}_{\mathrm{lum}}(\hat{\myvector{R}}_1)
        + \mathcal{L}_{\mathrm{lum}}(\hat{\myvector{R}}_2) \right) \nonumber \\
      &+ \lambda_5 \left( \mathcal{L}_{\ell_1}(\myvector{c}_1, \hat{\myvector{c}}_1)
        + \mathcal{L}_{\ell_1}(\myvector{c}_2, \hat{\myvector{c}}_2) \right), \\
    \mathcal{L}_{\mathrm{lum}}(\myvector{R})
      &= \frac{1}{|P|} \sum_{(x, y) \in P} \| 0.5 - \ip{\myvector{R}(x, y)}{\myvector{v}} \|_1
    \label{eq:r_loss}
  \end{align}
  where vector $\myvector{v} = (0.2126, 0.7152, 0.0722)^\top$ consists of coefficients
  for calculating the luminance of an RGB vector,
  and $\mathcal{L}_{\mathrm{lum}}$ is utilized for normalizing the luminance of reflectance to $0.5$.
  By normalizing the luminance to $0.5$, 
  our network can output reflectance $\hat{\myvector{R}}_i$
  so that it is consistent regardless of the illumination-brightness conditions.
  Moreover, by minimizing $\ell_1$-distance between $\myvector{c}_i$ and $\hat{\myvector{c}}_i$,
  reflectance $\hat{\myvector{R}}$ can also be consistent
  regardless of illumination-color conditions.

  $\lambda_1, \lambda_2, \lambda_3, \lambda_4$, and $\lambda_5$ are weights of the loss terms,
  In practice,
  we empirically set $\lambda_1 = 3, \lambda_2 = 1, \lambda_3 = 2, \lambda_4 = 1$,
  and $\lambda_5 = 1$ as weights.

\subsection{Network architecture}
\begin{figure*}[t]
    \centering
    \includegraphics[clip, width=0.95\hsize]{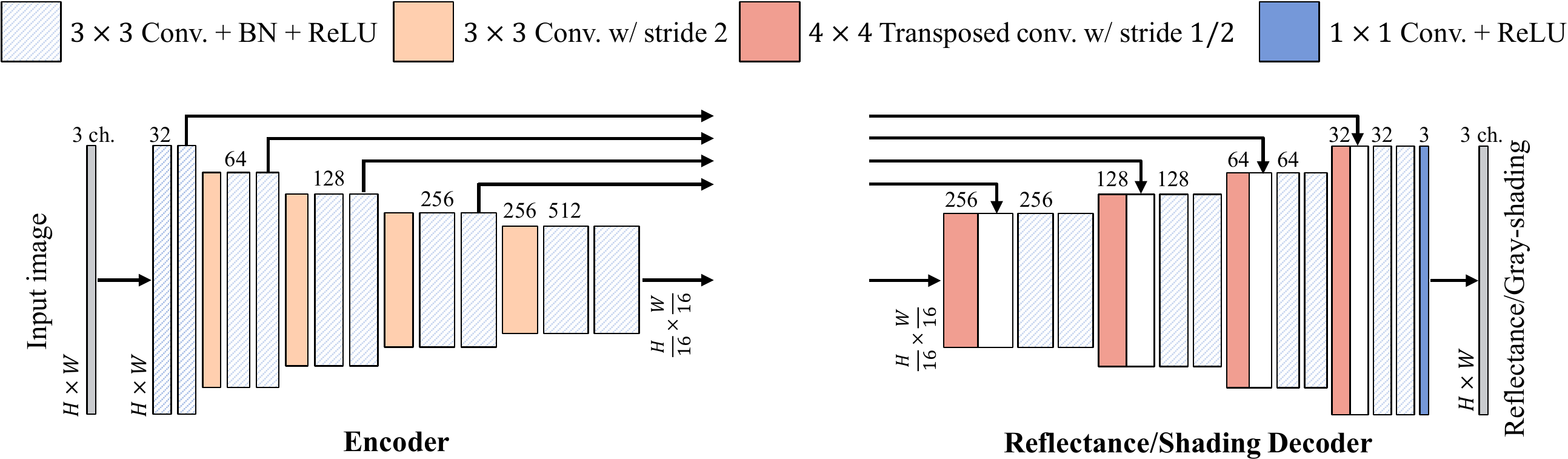}
    \caption{Network architecture of encoder and reflectance/shading decoders.
      Each box denotes multi-channel feature map produced by each layer.
      Number of channels is denoted above each box.
      Feature map resolutions are denoted to left of boxes.
      \label{fig:network}}
  \end{figure*}

  The proposed network consists of an encoder, a reflectance decoder, a shading decoder,
  and an illuminant-color decoder.
  Figure \ref{fig:network} shows the network architecture of
  the encoder and reflectance/shading decoders in detail.
  The input image is an RGB color image of $H \times W$ pixels.
  For the encoder and the reflectance/shading decoders,
  we use four types of layers as shown in Fig. \ref{fig:network}:
  \begin{description}[style=standard, nosep]
    \item [$3 \times 3$ Conv. + BN + ReLU],
      ~which calculates a $3 \times 3$ convolution
      with a stride of $1$ and a padding of $1$.
      After the convolution, batch normalization \cite{ioffe2015batch}
      and the rectified linear unit activation function \cite{glorot2011deep} (ReLU)
      are applied.
      Two adjacent $3 \times 3$ Conv. + BN + ReLU layers will have the same number $K$ of filters,
      where $K$ (i.e., the number of channels) is denoted in Fig. \ref{fig:network}.
    \item [$1 \times 1$ Conv. + ReLU],
      ~which calculates a $1 \times 1$ convolution with a stride of $1$ and without padding.
      After the convolution, ReLU is applied.
      The number of filters in the layer is $3$.
    \item [$3 \times 3$ Conv. w/ stride $2$],
      ~which downsamples feature maps using a $3 \times 3$ convolution
      with a stride of $2$ and a padding of $1$.
    \item [$4 \times 4$ Transposed conv. w/ stride $1/2$],
      ~which upsamples feature maps using a $4 \times 4$ transposed convolution
      with a stride of $1/2$ and a padding of $1$.
  \end{description}
  The proposed network has concatenated skip connections between the encoder and the decoders
  like U-Net~\cite{ronneberger2015unet},
  Feature maps from intermediate layers in the encoder
  are concatenated with those from intermediate layers in the reflectance and shading decoders.

  In addition to these three sub-networks,
  we utilize an illumination-color decoder that estimates the illumination color of a given image.
  The decoder consists of double $3 \times 3$ Conv. + BN + ReLU layers,
  a global average pooling layer, and a $1 \times 1$ conv. + ReLU layer.
  The $3 \times 3$ Conv. + BN + ReLU layers all have $64$ filters,
  and the $1 \times 1$ conv. + ReLU layer has $3$ filters.
  As a result, we can obtain a $3$-dimensional RGB vector that means the illumination color
  for each input image.

\section{Training proposed network in self-supervised manner}
\label{sec:self_supervised}
  To calculate the loss function for the proposed network,
  images of a single scene taken under various illumination-brightness and -color conditions are required.
  However, it is very costly to collect such images.
  For this reason, we generate pseudo images from raw images
  and use them for training the proposed network.

\subsection{Simulating brightness change}
  A change in illumination brightness, i.e.,
  radiance, affects image brightness.
  Here, we assume that camera parameters except for the shutter speed are fixed,
  although the image brightness can be controlled by adjusting various camera parameters.
  In this case,
  the image brightness is determined by the radiant power density at an imaging sensor,
  i.e., irradiance, and shutter speed,
  where the irradiance is proportional to the radiance.

  Under the assumption that a scene is static during the time that the shutter is open
  and the sensor has a linear response with respect to the light intensity,
  the image brightness is proportional to the radiance (and also the shutter speed)~\cite{kinoshita2019scene}.
  Hence, the change in illumination brightness is simulated by multiplying an image $\myvector{I}$
  by a scalar value of $2^v$ as
  \begin{equation}
    \myvector{I}'(x, y) = 2^{v} \myvector{I}(x, y),
    \label{eq:brightness}
  \end{equation}
  where it is equivalent to the exposure compensation of $v [\mathrm{EV}]$~\cite{kinoshita2019scene}.

\subsection{Simulating color change}
  The effect of illumination-color change is already considered
  in the research area of white balancing.
  White balancing aims to remove the effects of illumination color on an image.
  Typically, white balancing is performed by
  \begin{equation}
    \myvector{I}''(x, y) = \mymatrix{M}_{\mathrm{WB}} \myvector{I}(x, y),
    \label{eq:wb}
  \end{equation}
  where
  \begin{equation}
    \mymatrix{M}_{\mathrm{WB}} =
    \mymatrix{M}_A^{-1}
    \begin{pmatrix}
      \alpha & 0 & 0 \\
      0 & \beta & 0 \\
      0 & 0 & \gamma \\
    \end{pmatrix}
    \mymatrix{M}_A
    \label{eq:wb_matrix}
  \end{equation}
  and $\mymatrix{M}_A$ is a full-rank matrix with a size of $3 \times 3$ for color space conversion.
  Parameters $\alpha, \beta$, and $\gamma$ are non-negative, and they are calculated
  from the ideal white illumination and illumination color $\myvector{c}$ of an input image.

  By using the inverse of $\mymatrix{M}_{\mathrm{WB}}$,
  we can perform \textit{inverse} white balancing as
  \begin{equation}
    \myvector{I}(x, y) = \mymatrix{M}_{\myvector{c}} \myvector{I}''(x, y),
    \label{eq:color}
  \end{equation}
  where $\mymatrix{M}_{\myvector{c}} = \mymatrix{M}_{\mathrm{WB}}^{-1}$.
  This corresponds to simulating illumination-color change.

  For these reasons, illumination-brightness and -color changes can be simulated
  in accordance with Eqs. (\ref{eq:brightness}) and (\ref{eq:wb}), respectively,
  by using various $v, \alpha, \beta$, and $\gamma$.
  Images generated by using the equations can be used for training the proposed network.

\subsection{Procedure for data generation \label{subsec:procedure}}
  In accordance with Eqs. (\ref{eq:brightness}) and (\ref{eq:color}),
  we randomly simulate images of a single scene taken under various illuminant-brightness and -color conditions,
  at every step of training.
  Images $\myvector{I}_i$ are generated
  from a raw image $\myvector{I}_{\mathrm{raw}}$ as follows.
  \begin{enumerate}[leftmargin=*]
    \item Obtain an RGB image $\myvector{I}$
      by demosaicing and white-balancing a raw image $\myvector{I}_{\mathrm{raw}}$.
    \item Simulate illumination-color change in accordance with Eq.(\ref{eq:brightness}) as
      \begin{equation}
        \myvector{I}'_i(x, y) = 2^{v_i}\frac{0.18}{g(\myvector{I})} \myvector{I}(x, y),
        \label{eq:scala}
      \end{equation}
      where $v_i$ is a uniform random number in the range of $[-1, 1]$, and 
      $g(\myvector{I})$ indicates
      the geometric mean of the luminance of $\myvector{I}$~\cite{reinhard2002photographic}.
    \item Simulate illumination-color change in accordance with Eq.(\ref{eq:color})
      by multiplying $\myvector{I}'_i$ by $\mymatrix{M}_{c_i}$ as
      \begin{equation}
        \myvector{I}_i(x, y) = \mymatrix{M}_{c_i} \myvector{I}'_i(x, y),
        \label{eq:ctrans}
      \end{equation}
      where $\mymatrix{M}_{\myvector{c}_i}$ is a diagonal matrix having $\myvector{c}_i$ as the main diagonal,
      and $\myvector{c}_i$ is a uniform random vector in $[0.9, 1.1]^3$.
  \end{enumerate}

\section{Simulation} \label{sec:simulation}
  We performed two simulation experiments to evaluate the proposed network
  in terms of the three consistencies, i.e.,
  reconstruction consistency,
  reflectance consistency (brightness),
  and reflectance consistency (color).

\subsection{Simulation using raw images \label{subsec:raw_simulation}}
\subsubsection{Experimental conditions}
\begin{figure}[t]
      \centering
  	  \begin{subfigure}[t]{0.3\hsize}
        \centering
  		  \includegraphics[width=\columnwidth]{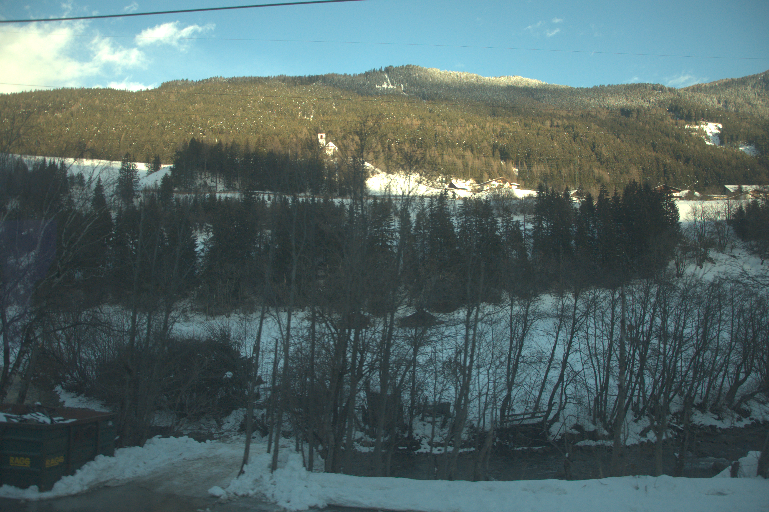}
        \caption{$\myvector{I}_1$}
      \end{subfigure}
      \begin{subfigure}[t]{0.3\hsize}
        \centering
  		  \includegraphics[width=\columnwidth]{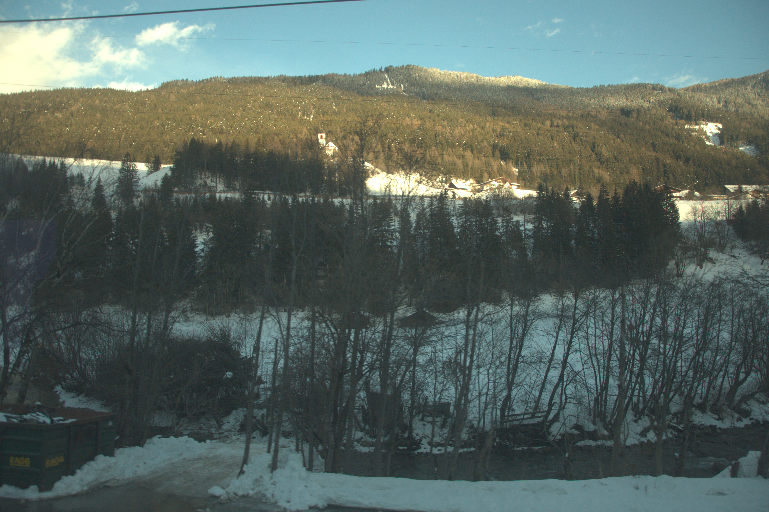}
        \caption{$\myvector{I}_2$}
      \end{subfigure}
  	  \begin{subfigure}[t]{0.3\hsize}
        \centering
  		  \includegraphics[width=\columnwidth]{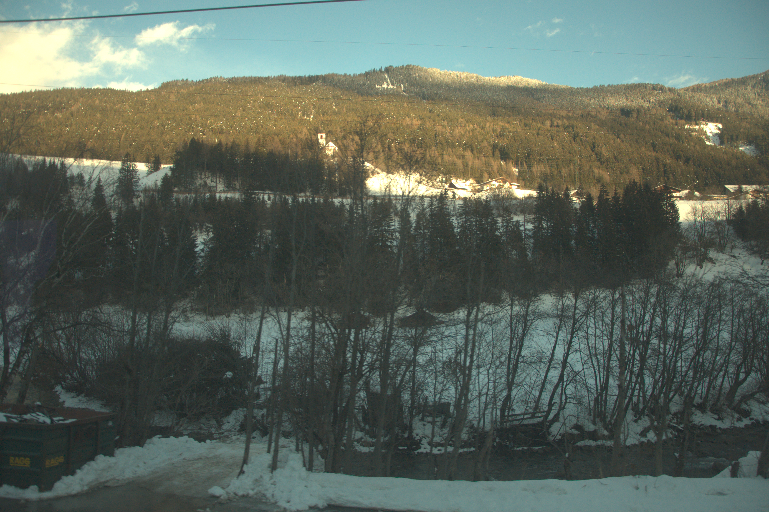}
        \caption{$\myvector{I}_3$}
      \end{subfigure}\\
  	  \begin{subfigure}[t]{0.3\hsize}
        \centering
  		  \includegraphics[width=\columnwidth]{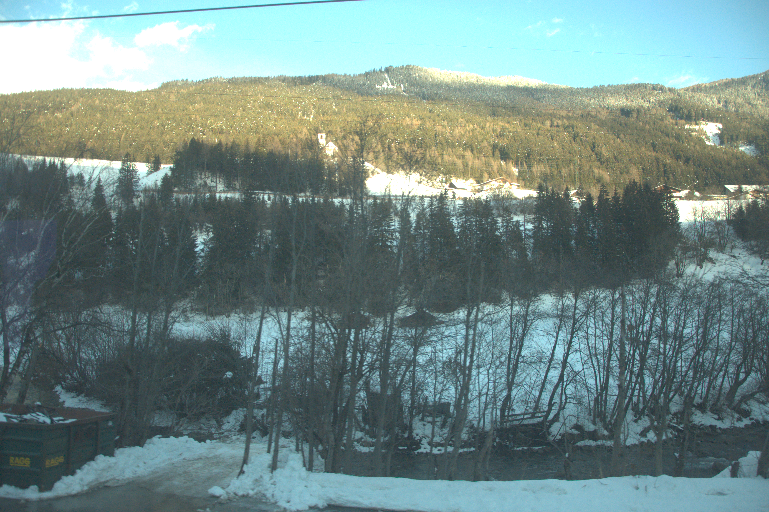}
        \caption{$\myvector{I}_4$}
      \end{subfigure}
      \begin{subfigure}[t]{0.3\hsize}
        \centering
  		  \includegraphics[width=\columnwidth]{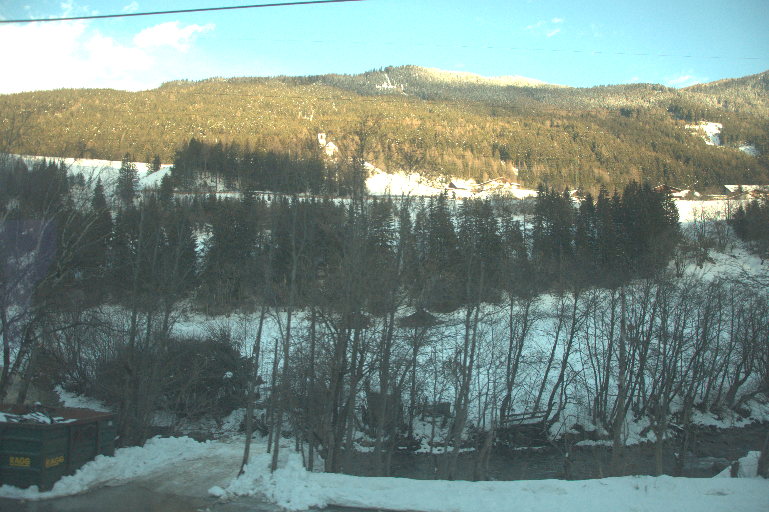}
        \caption{$\myvector{I}_5$}
      \end{subfigure}
  	  \begin{subfigure}[t]{0.3\hsize}
        \centering
  		  \includegraphics[width=\columnwidth]{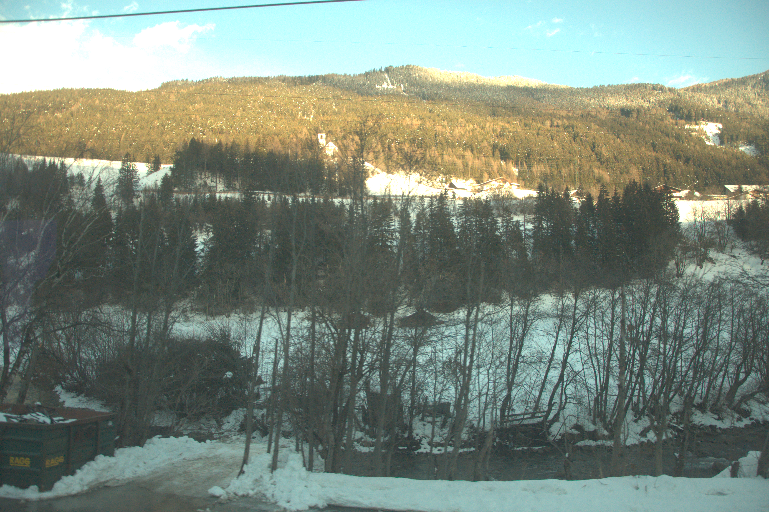}
        \caption{$\myvector{I}_6$}
      \end{subfigure}\\
  	  \begin{subfigure}[t]{0.3\hsize}
        \centering
  		  \includegraphics[width=\columnwidth]{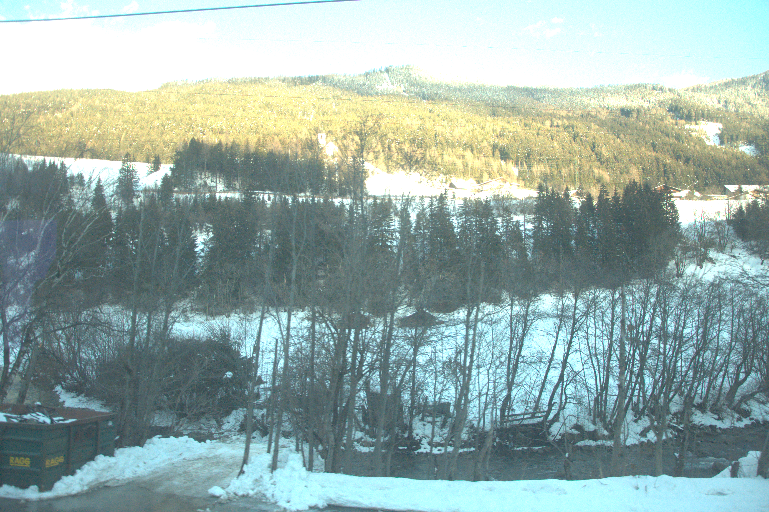}
        \caption{$\myvector{I}_7$}
      \end{subfigure}
      \begin{subfigure}[t]{0.3\hsize}
        \centering
  		  \includegraphics[width=\columnwidth]{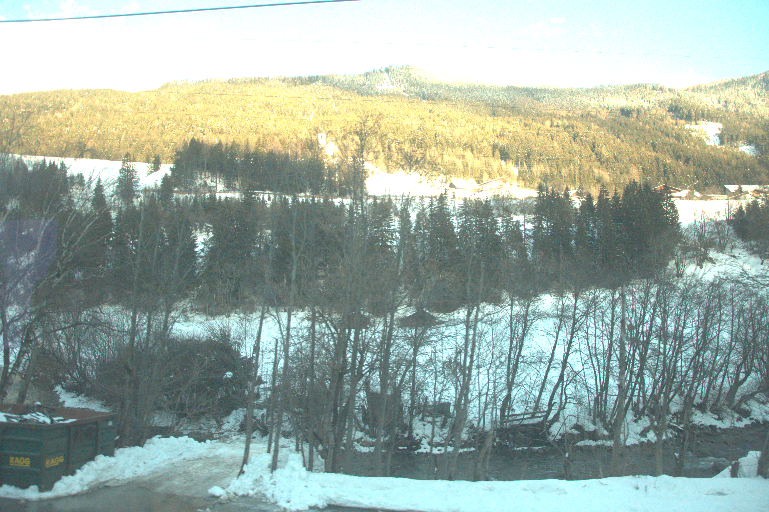}
        \caption{$\myvector{I}_8$}
      \end{subfigure}
  	  \begin{subfigure}[t]{0.3\hsize}
        \centering
  		  \includegraphics[width=\columnwidth]{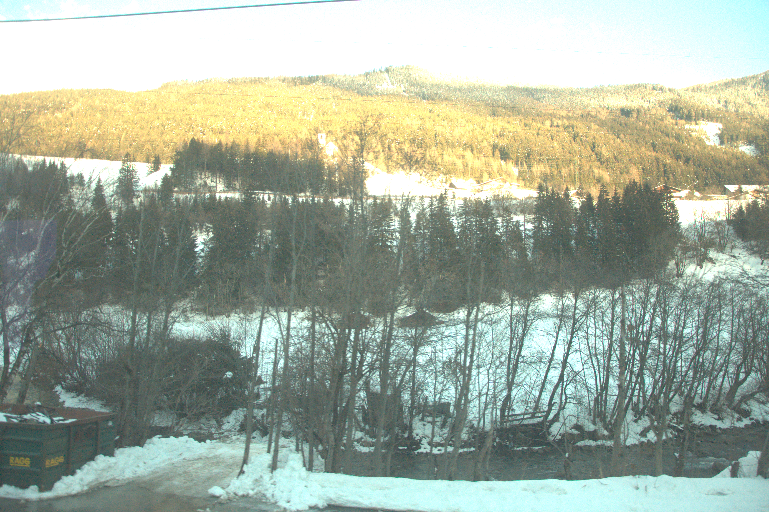}
        \caption{$\myvector{I}_9$}
      \end{subfigure}
      \caption{Example of input image set.
        Row denotes illumination brightness [Top: -1 EV, Middle: 0 EV, and Bottom: +1 EV],
        and column denotes illumination color
        [Left: Cold white (0.9, 1.0, 1.1), Middle: White (1.0, 1.0, 1.0), and Right: Warm white (1.1, 1.0, 0.9)]
        \label{fig:inputs}}
  \end{figure}

  In this experiment,
  45 image sets were generated from 45 raw images from the RAISE Dataset \cite{nguyen2015raise},
  where each image set consists of $9$ images.
  Figure \ref{fig:inputs} illustrates an example of an input image set.
  The generation procedure was the same as that in Section \ref{sec:self_supervised}.\ref{subsec:procedure}),
  where $v_i \in \{-1, 0, +1\}$, and $\myvector{c}_i \in \{(0.9, 1.0, 1.1), (1.0, 1.0, 1.0), (1.1, 1.0, 0.9)\}$.
  Each generated image $\myvector{I}_i$ was independently fed into the trained network
  and decomposed into gray-shading $\hat{S}_i$, reflectance $\hat{\myvector{R}}_i$,
  and illumination color $\hat{\myvector{c}}_i$.

  To evaluate the reconstruction consistency,
  we used the peak signal-to-noise ratio (PSNR), mean squared error (MSE), and structural dissimilarity (DSSIM)
  between input image $\myvector{I}_i$ and reconstructed image $\hat{\myvector{I}}_i$.
  Here, a larger PSNR value means higher similarity between two images.
  For MSE and DSSIM, a smaller value means higher similarity.
  To evaluate the reflectance consistency in terms of both brightness and color,
  we also used the PSNR, MSE, and DSSIM between a reference reflectance
  and an estimated reflectance $\hat{\myvector{R}}_i$ of an image under another illumination condition.
  Here, $\hat{\myvector{R}}_5$ was used as the reference reflectance.

  The proposed network was compared with the following methods:
  \begin{itemize}
    \item Bell's decomposition method~\cite{bell2014intrinsic},
    \item Unsupervised Learning for Intrinsic Image Decomposition from a Single Image (USI3D)~\cite{liu2020unsupervised},
  \end{itemize}
  where Bell's method is an optimization based method
  and USI3D is a state-of-the-art DNN-based method.
  We used the authors' original implementations of the two methods, which are available on GitHub.

  Our network was trained with 100 epochs
  by using 3620 raw images from the HDR+ burst photography dataset~\cite{hasinoff2016burst}.
  For data augmentation,
  we resized each original input image with a random scaling factor in the range of $[0.6, 1.0]$
  for every epoch.
  After the resizing, we randomly cropped the resized image to
  an image patch with a size of $256 \times 256$ pixels
  and flipped the patch horizontally with a probability of 0.5.
  In addition to the augmentation,
  we generated two images by applying
  the procedure in Section \ref{sec:self_supervised}.\ref{subsec:procedure})
  to each augmented image
  in order to simulate various illumination-brightness and -color conditions.
  Loss was calculated by using the two generated images
  in accordance with Eq. (\ref{eq:all_loss}) to Eq. (\ref{eq:r_loss}).
  Here, the Adam optimizer \cite{kingma2014adam} was utilized for optimization,
  where the parameters in Adam were set as $\alpha=0.001, \beta_1=0.9$, and $\beta_2=0.999$.
  He's method \cite{he2015delving} was used for initializing the network.

\subsubsection{Experimental results}
\begin{figure*}[t]
    \centering
    \begin{subfigure}[t]{0.3\hsize}
      \centering
  		\includegraphics[width=\columnwidth]{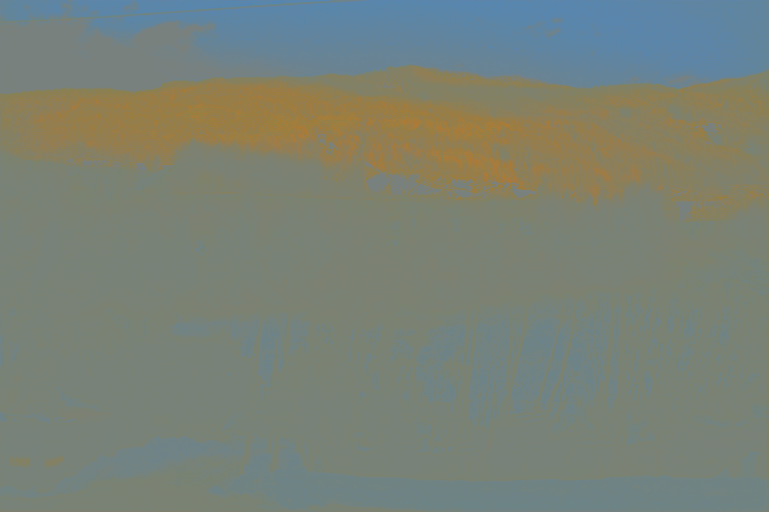}
      \caption{Reflectance $\hat{\myvector{R}}$}
    \end{subfigure}
    \begin{subfigure}[t]{0.3\hsize}
      \centering
  		\includegraphics[width=\columnwidth]{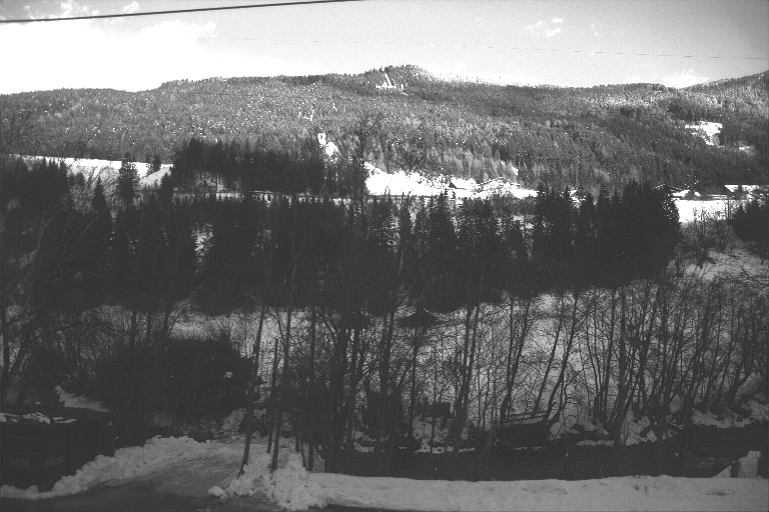}
      \caption{Gray shading $\hat{S}$}
    \end{subfigure}
    \begin{subfigure}[t]{0.3\hsize}
      \centering
  		\includegraphics[width=\columnwidth]{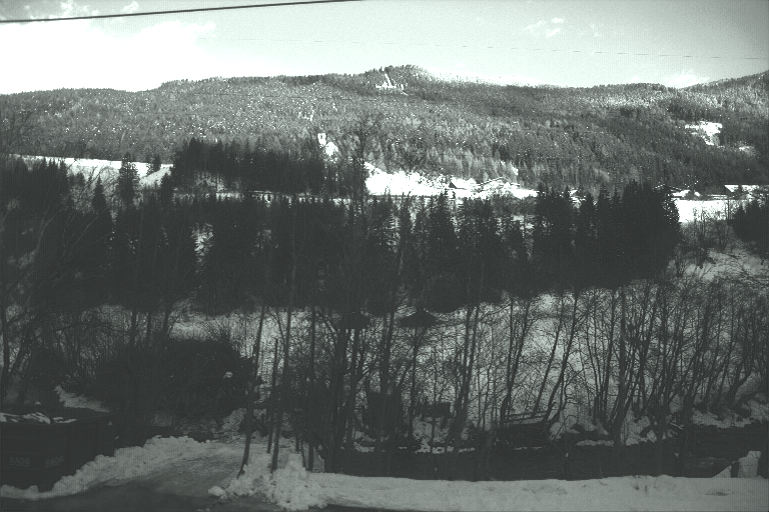}
      \caption{Shading $\hat{\myvector{S}}$}
    \end{subfigure}
    \caption{Example of decomposition results by our network.
      Input image was $\myvector{I}_6$ in Fig. \ref{fig:inputs}.
    \label{fig:result}}
  \end{figure*}

  Figure \ref{fig:result} shows an example of decomposition with our network.
  From Fig. \ref{fig:result}, we can see that the brightness of estimated reflectance $\hat{\myvector{R}}$
  was constant,
  and the estimated shading $\hat{\myvector{S}}$ had a single color.
  
\begin{figure}[t]
      \centering
  	  \begin{subfigure}[t]{0.3\hsize}
        \centering
  		  \includegraphics[width=\columnwidth]{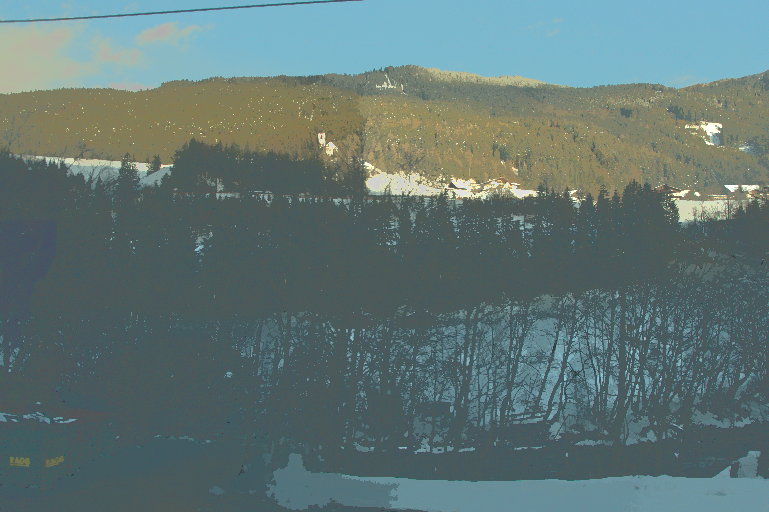}
        \caption{$\hat{\myvector{R}}_1$}
      \end{subfigure}
      \begin{subfigure}[t]{0.3\hsize}
        \centering
  		  \includegraphics[width=\columnwidth]{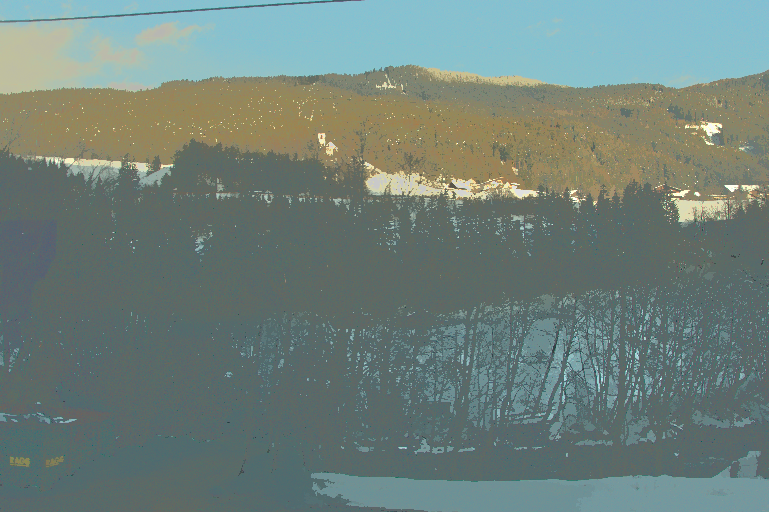}
        \caption{$\hat{\myvector{R}}_2$}
      \end{subfigure}
  	  \begin{subfigure}[t]{0.3\hsize}
        \centering
  		  \includegraphics[width=\columnwidth]{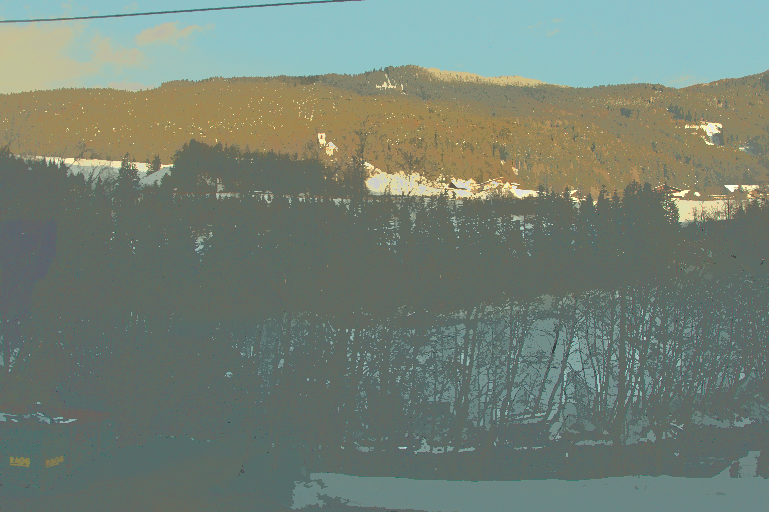}
        \caption{$\hat{\myvector{R}}_3$}
      \end{subfigure}\\
  	  \begin{subfigure}[t]{0.3\hsize}
        \centering
  		  \includegraphics[width=\columnwidth]{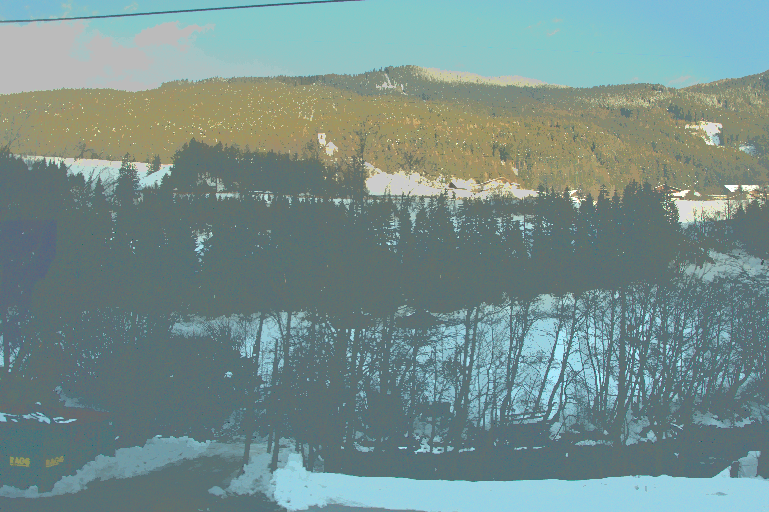}
        \caption{$\hat{\myvector{R}}_4$}
      \end{subfigure}
      \begin{subfigure}[t]{0.3\hsize}
        \centering
  		  \includegraphics[width=\columnwidth]{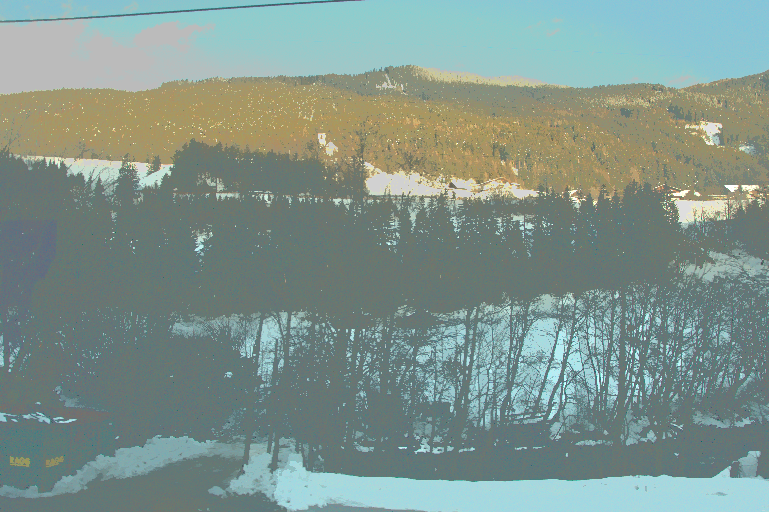}
        \caption{$\hat{\myvector{R}}_5$}
      \end{subfigure}
  	  \begin{subfigure}[t]{0.3\hsize}
        \centering
  		  \includegraphics[width=\columnwidth]{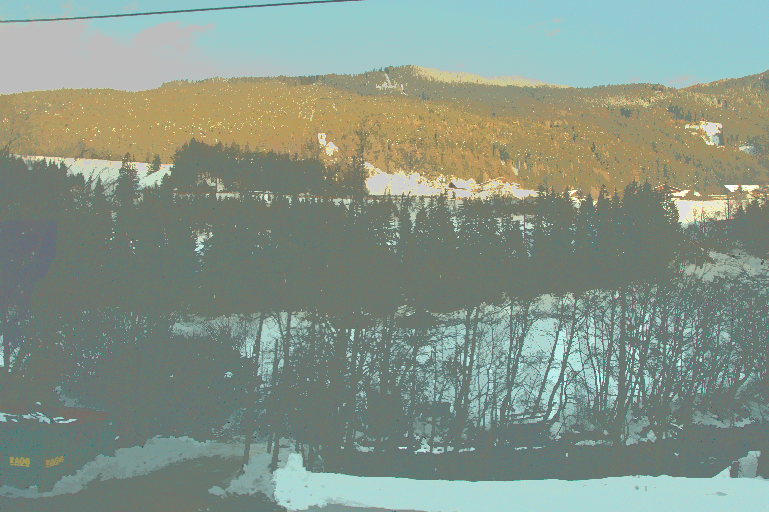}
        \caption{$\hat{\myvector{R}}_6$}
      \end{subfigure}\\
  	  \begin{subfigure}[t]{0.3\hsize}
        \centering
  		  \includegraphics[width=\columnwidth]{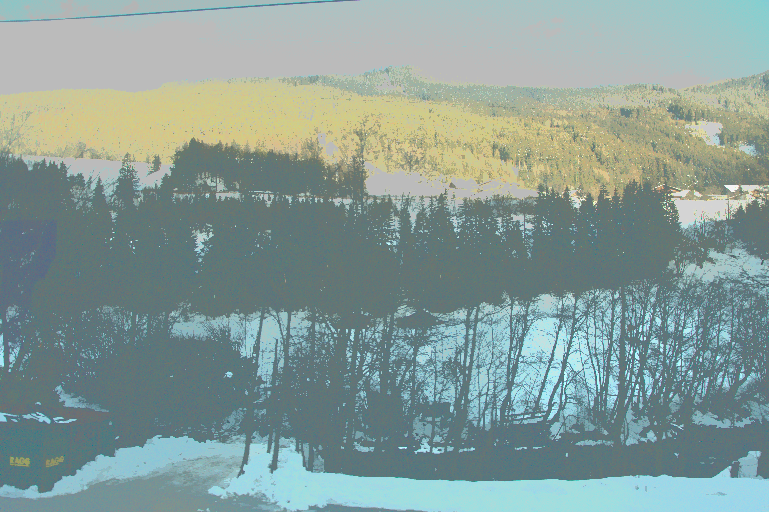}
        \caption{$\hat{\myvector{R}}_7$}
      \end{subfigure}
      \begin{subfigure}[t]{0.3\hsize}
        \centering
  		  \includegraphics[width=\columnwidth]{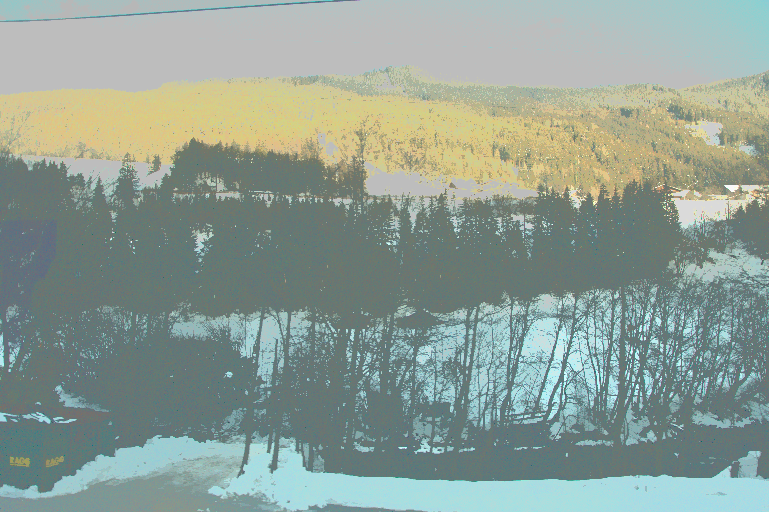}
        \caption{$\hat{\myvector{R}}_8$}
      \end{subfigure}
  	  \begin{subfigure}[t]{0.3\hsize}
        \centering
  		  \includegraphics[width=\columnwidth]{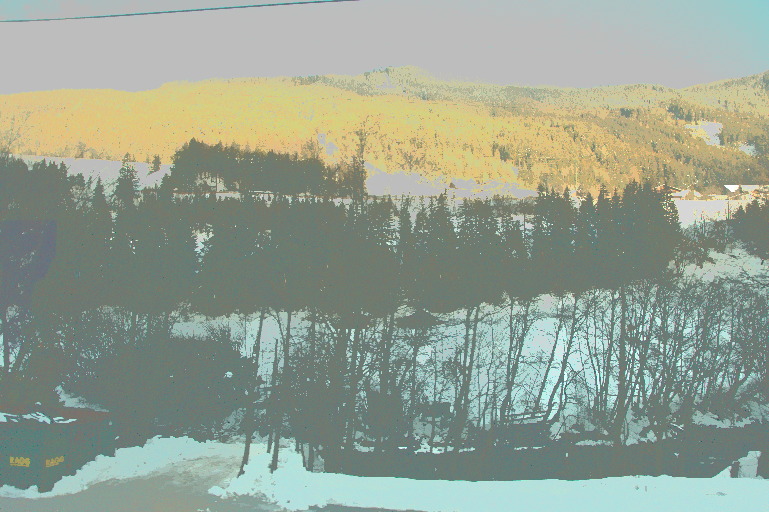}
        \caption{$\hat{\myvector{R}}_9$}
      \end{subfigure}
      \caption{Example of estimated reflectance for images in Fig. \ref{fig:inputs} (Bell's method~\cite{bell2014intrinsic}).
        Row denotes illumination brightness [Top: -1 EV, Middle: 0 EV, and Bottom: +1 EV],
        and column denotes illumination color
        [Left: Cold white (0.9, 1.0, 1.1), Middle: White (1.0, 1.0, 1.0), and Right: Warm white (1.1, 1.0, 0.9)]
        \label{fig:bell_r}}
  \end{figure}

\begin{figure}[t]
      \centering
  	  \begin{subfigure}[t]{0.3\hsize}
        \centering
  		  \includegraphics[width=\columnwidth]{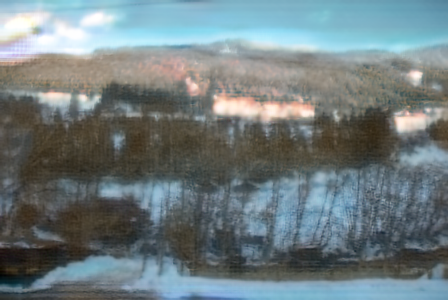}
        \caption{$\hat{\myvector{R}}_1$}
      \end{subfigure}
      \begin{subfigure}[t]{0.3\hsize}
        \centering
  		  \includegraphics[width=\columnwidth]{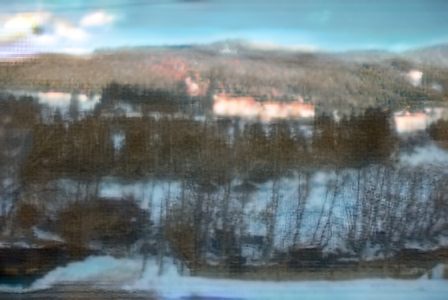}
        \caption{$\hat{\myvector{R}}_2$}
      \end{subfigure}
  	  \begin{subfigure}[t]{0.3\hsize}
        \centering
  		  \includegraphics[width=\columnwidth]{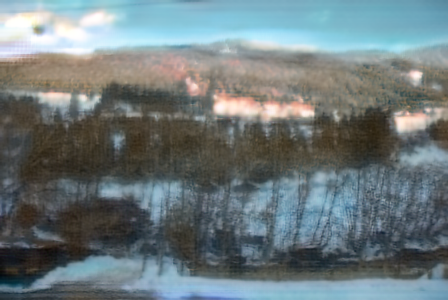}
        \caption{$\hat{\myvector{R}}_3$}
      \end{subfigure}\\
  	  \begin{subfigure}[t]{0.3\hsize}
        \centering
  		  \includegraphics[width=\columnwidth]{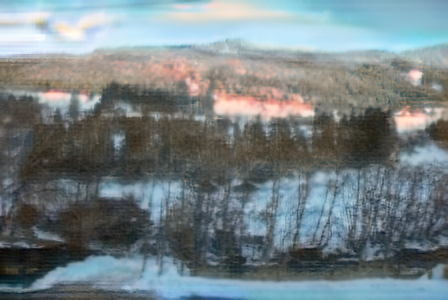}
        \caption{$\hat{\myvector{R}}_4$}
      \end{subfigure}
      \begin{subfigure}[t]{0.3\hsize}
        \centering
  		  \includegraphics[width=\columnwidth]{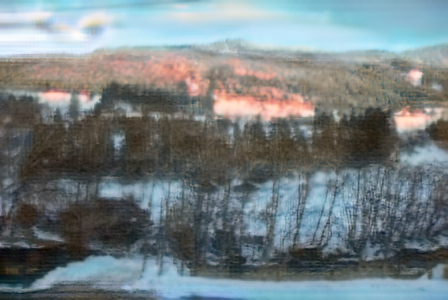}
        \caption{$\hat{\myvector{R}}_5$}
      \end{subfigure}
  	  \begin{subfigure}[t]{0.3\hsize}
        \centering
  		  \includegraphics[width=\columnwidth]{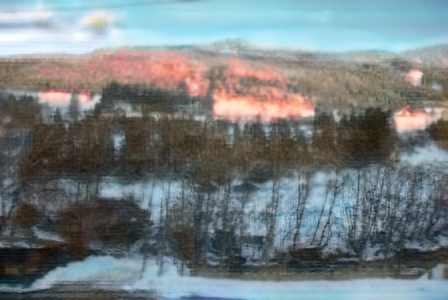}
        \caption{$\hat{\myvector{R}}_6$}
      \end{subfigure}\\
  	  \begin{subfigure}[t]{0.3\hsize}
        \centering
  		  \includegraphics[width=\columnwidth]{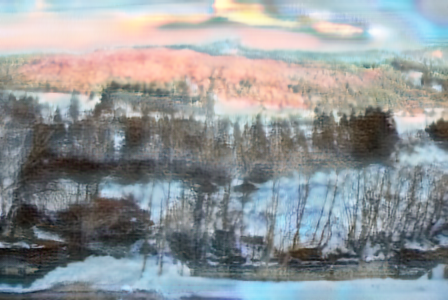}
        \caption{$\hat{\myvector{R}}_7$}
      \end{subfigure}
      \begin{subfigure}[t]{0.3\hsize}
        \centering
  		  \includegraphics[width=\columnwidth]{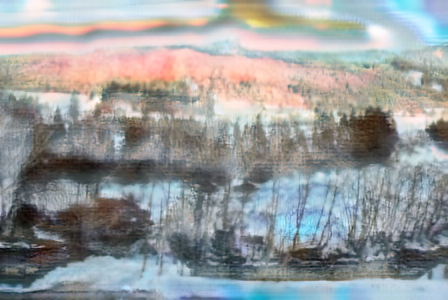}
        \caption{$\hat{\myvector{R}}_8$}
      \end{subfigure}
  	  \begin{subfigure}[t]{0.3\hsize}
        \centering
  		  \includegraphics[width=\columnwidth]{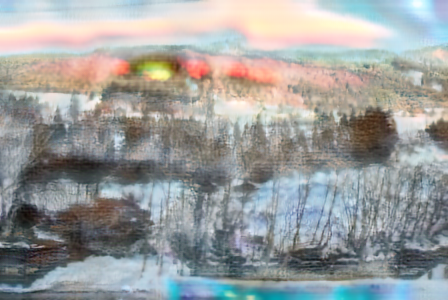}
        \caption{$\hat{\myvector{R}}_9$}
      \end{subfigure}
      \caption{Example of estimated reflectance for images in Fig. \ref{fig:inputs} (USI3D~\cite{liu2020unsupervised}).
        Row denotes illumination brightness [Top: -1 EV, Middle: 0 EV, and Bottom: +1 EV],
        and column denotes illumination color
        [Left: Cold white (0.9, 1.0, 1.1), Middle: White (1.0, 1.0, 1.0), and Right: Warm white (1.1, 1.0, 0.9)]
        \label{fig:usi3d_r}}
  \end{figure}

\begin{figure}[t]
      \centering
  	  \begin{subfigure}[t]{0.3\hsize}
        \centering
  		  \includegraphics[width=\columnwidth]{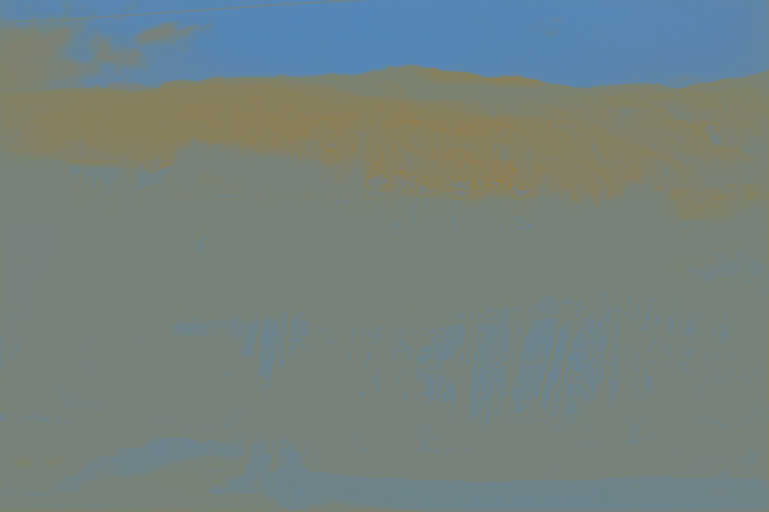}
        \caption{$\hat{\myvector{R}}_1$}
      \end{subfigure}
      \begin{subfigure}[t]{0.3\hsize}
        \centering
  		  \includegraphics[width=\columnwidth]{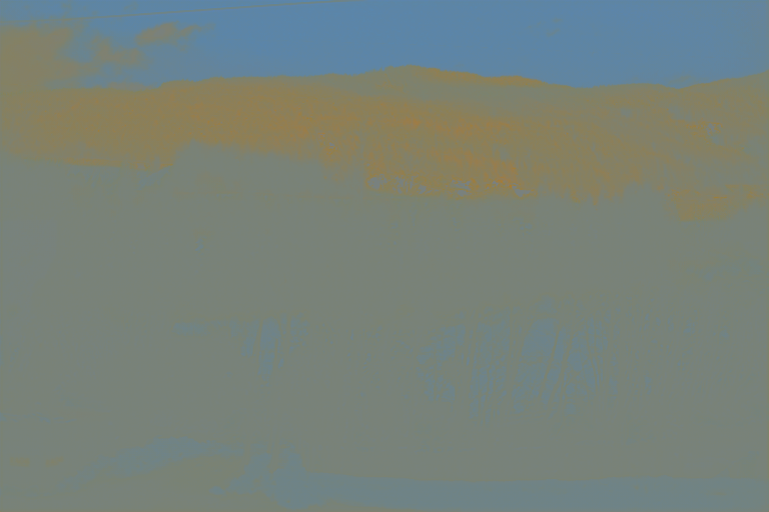}
        \caption{$\hat{\myvector{R}}_2$}
      \end{subfigure}
  	  \begin{subfigure}[t]{0.3\hsize}
        \centering
  		  \includegraphics[width=\columnwidth]{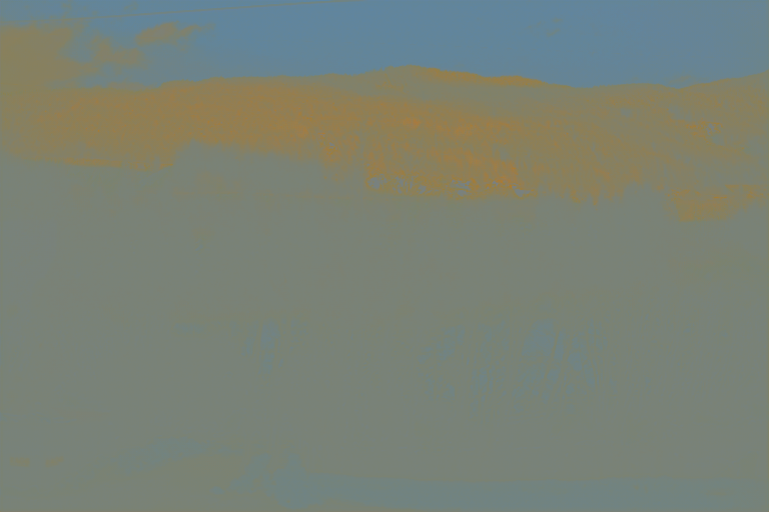}
        \caption{$\hat{\myvector{R}}_3$}
      \end{subfigure}\\
  	  \begin{subfigure}[t]{0.3\hsize}
        \centering
  		  \includegraphics[width=\columnwidth]{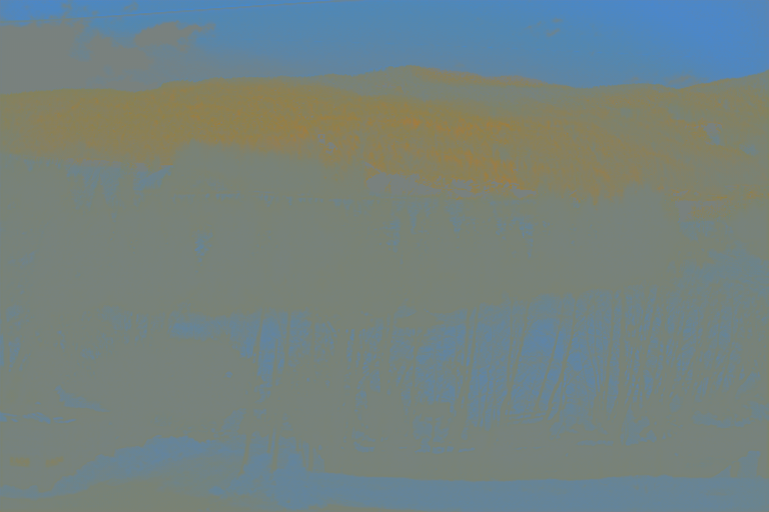}
        \caption{$\hat{\myvector{R}}_4$}
      \end{subfigure}
      \begin{subfigure}[t]{0.3\hsize}
        \centering
  		  \includegraphics[width=\columnwidth]{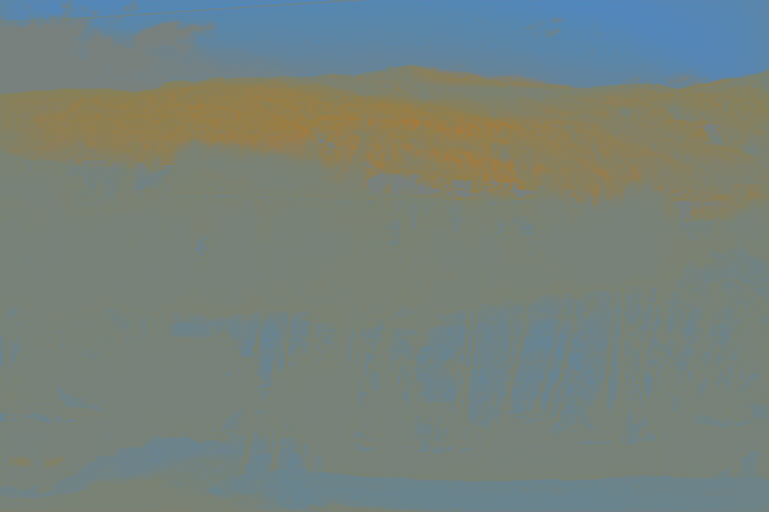}
        \caption{$\hat{\myvector{R}}_5$}
      \end{subfigure}
  	  \begin{subfigure}[t]{0.3\hsize}
        \centering
  		  \includegraphics[width=\columnwidth]{figs/proposed/r00b3931bt__0ev_1.1__1_0.9_reflectance.png}
        \caption{$\hat{\myvector{R}}_6$}
      \end{subfigure}\\
  	  \begin{subfigure}[t]{0.3\hsize}
        \centering
  		  \includegraphics[width=\columnwidth]{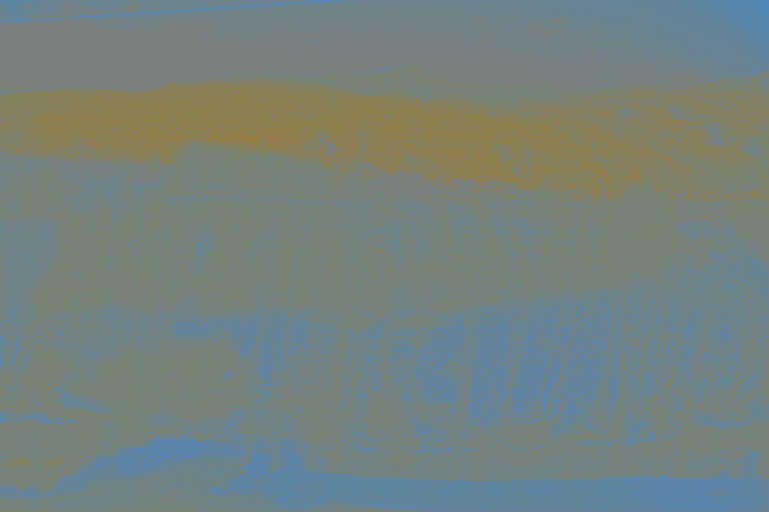}
        \caption{$\hat{\myvector{R}}_7$}
      \end{subfigure}
      \begin{subfigure}[t]{0.3\hsize}
        \centering
  		  \includegraphics[width=\columnwidth]{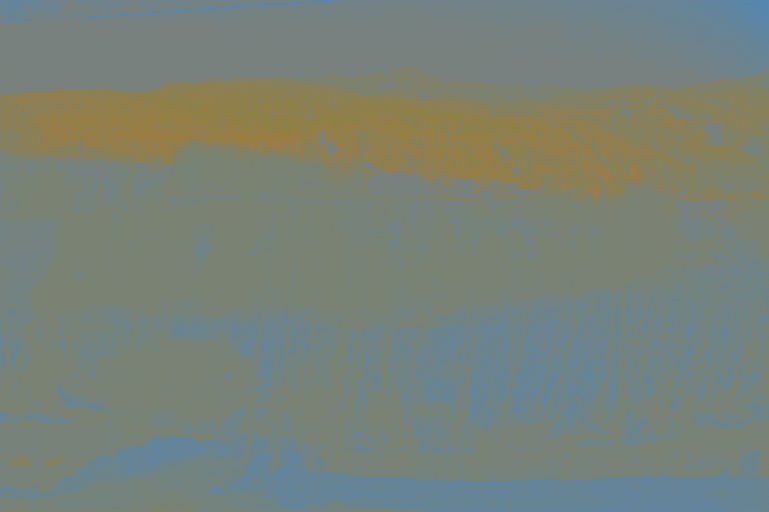}
        \caption{$\hat{\myvector{R}}_8$}
      \end{subfigure}
  	  \begin{subfigure}[t]{0.3\hsize}
        \centering
  		  \includegraphics[width=\columnwidth]{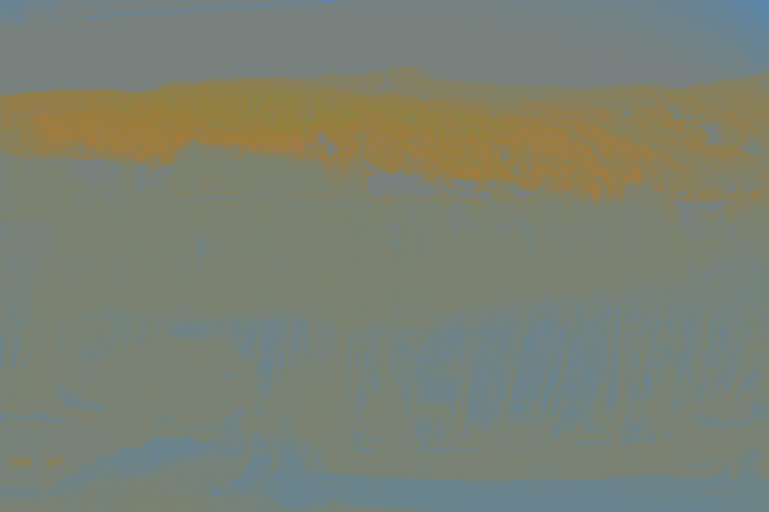}
        \caption{$\hat{\myvector{R}}_9$}
      \end{subfigure}
      \caption{Example of estimated reflectance for images in Fig. \ref{fig:inputs} (Proposed).
        Row denotes illumination brightness [Top: -1 EV, Middle: 0 EV, and Bottom: +1 EV],
        and column denotes illumination color
        [Left: Cold white (0.9, 1.0, 1.1), Middle: White (1.0, 1.0, 1.0), and Right: Warm white (1.1, 1.0, 0.9)]
        \label{fig:proposed_r}}
  \end{figure}

  Figures \ref{fig:bell_r}, \ref{fig:usi3d_r}, and \ref{fig:proposed_r} are
  references of the images in Fig. \ref{fig:inputs} estimated by Bell's method, USI3D, and the proposed network,
  respectively.
  From these figures, we can see that the estimated reflectance by the three methods is different.
  This is because the proposed network takes into account not only conventional reconstruction consistency
  but also two reflectance consistencies.
  The two reflectance consistencies require estimated reflectance
  to satisfy that it is independent from lighting conditions, i.e.,
  brightness and illumination color.
  As shown in Figs \ref{fig:bell_r} and \ref{fig:usi3d_r},
  the conventional decomposition methods produced reflectance that depended on illumination brightness and color
  owing to the lack of consideration of reflectance consistencies.
  In contrast,
  the proposed network produced almost the same reflectance for all nine images
  that depended on illumination brightness and color.
  For this reason, the proposed network satisfied the reflectance consistencies in terms of brightness
  and color, but the conventional methods did not.

  The reflectance consistencies of the proposed network were also confirmed in a numerical evaluation.
  Tables \ref{tab:reflect}, \ref{tab:reflect_mse}, and \ref{tab:reflect_dssim}
  show the PSNR, MSE, and DSSIM values between $\hat{\myvector{R}}_i$ and $\hat{\myvector{R}}_5$
  for each $i$ for the three methods, where these values were averaged over the 45 image sets
  that we used in this simulation.
  From Table \ref{tab:reflect}, the proposed network provided higher PSNR values (about $30 \mathrm{dB}$)
  than those of the conventional methods.
  Hence, the proposed network was shown to be robust against illumination-brightness/-color changes.
  This trend was also confirmed from Tables \ref{tab:reflect_mse} and \ref{tab:reflect_dssim}.
  \begin{table}[t]
    \centering
    \caption{PSNR values between $\hat{\myvector{R}}_i$ and $\hat{\myvector{R}}_5$
      averaged over 45 image sets.
      Top: PSNR between $\hat{\myvector{R}}_1$ and $\hat{\myvector{R}}_5$,
      Bottom: PSNR between $\hat{\myvector{R}}_9$ and $\hat{\myvector{R}}_5$}
    {\footnotesize
    \begin{tabular}{l|ccc} \hline\hline
                  & Bell~\cite{bell2014intrinsic} & USI3D~\cite{liu2020unsupervised} & Proposed \\ \hline
      $\hat{\myvector{R}}_1$ (-1 {[}EV{]}, Cold white) & 20.89 & 24.55 & 30.86 \\
      $\hat{\myvector{R}}_2$ (-1 {[}EV{]}, White)      & 21.62 & 25.10 & 31.00 \\
      $\hat{\myvector{R}}_3$ (-1 {[}EV{]}, Warm white) & 20.77 & 24.65 & 30.09 \\
      $\hat{\myvector{R}}_4$ (0 {[}EV{]}, Cold white)  & 26.09 & 33.58 & 34.40 \\
      $\hat{\myvector{R}}_5$ (0 {[}EV{]}, White)       & --    & --    & --    \\
      $\hat{\myvector{R}}_6$ (0 {[}EV{]}, Warm White)  & 26.75 & 32.97 & 34.56 \\
      $\hat{\myvector{R}}_7$ (+1 {[}EV{]}, Cold white) & 18.22 & 18.90 & 28.84 \\
      $\hat{\myvector{R}}_8$ (+1 {[}EV{]}, White)      & 18.41 & 18.31 & 30.59 \\
      $\hat{\myvector{R}}_9$ (+1 {[}EV{]}, Warm white) & 18.44 & 17.60 & 29.30 \\ \hline\hline
    \end{tabular}
    }
    \label{tab:reflect}
  \end{table}
  \begin{table}[t]
    \centering
    \caption{MSE values between $\hat{\myvector{R}}_i$ and $\hat{\myvector{R}}_5$
      averaged over 45 image sets.
      Top: MSE between $\hat{\myvector{R}}_1$ and $\hat{\myvector{R}}_5$,
      Bottom: MSE between $\hat{\myvector{R}}_9$ and $\hat{\myvector{R}}_5$}
    {\footnotesize
    \begin{tabular}{l|ccc} \hline\hline
                  & Bell~\cite{bell2014intrinsic} & USI3D~\cite{liu2020unsupervised} & Proposed \\ \hline
      $\hat{\myvector{R}}_1$ (-1 [EV], Cold white) & 0.013 & 0.007 & 0.001    \\
      $\hat{\myvector{R}}_2$ (-1 [EV], White)      & 0.010 & 0.007 & 0.001    \\
      $\hat{\myvector{R}}_3$ (-1 [EV], Warm white) & 0.011 & 0.007 & 0.001    \\
      $\hat{\myvector{R}}_4$ (0 [EV], Cold white)  & 0.004 & 0.001 & 0.000    \\
      $\hat{\myvector{R}}_5$ (0 [EV], White)       & 0.000 & 0.000 & 0.000    \\
      $\hat{\myvector{R}}_6$ (0 [EV], Warm white)  & 0.003 & 0.001 & 0.000    \\
      $\hat{\myvector{R}}_7$ (1 [EV], Cold white)  & 0.016 & 0.015 & 0.001    \\
      $\hat{\myvector{R}}_8$ (1 [EV], White)       & 0.017 & 0.018 & 0.001    \\
      $\hat{\myvector{R}}_9$ (1 [EV], Warm white)  & 0.017 & 0.020 & 0.001    \\ \hline\hline
    \end{tabular}
    }
    \label{tab:reflect_mse}
  \end{table}
  \begin{table}[t]
    \centering
    \caption{DSSIM values between $\hat{\myvector{R}}_i$ and $\hat{\myvector{R}}_5$
      averaged over 45 image sets.
      Top: DSSIM between $\hat{\myvector{R}}_1$ and $\hat{\myvector{R}}_5$,
      Bottom: DSSIM between $\hat{\myvector{R}}_9$ and $\hat{\myvector{R}}_5$}
    {\footnotesize
    \begin{tabular}{l|ccc} \hline\hline
                  & Bell~\cite{bell2014intrinsic} & USI3D~\cite{liu2020unsupervised} & Proposed \\ \hline
      $\hat{\myvector{R}}_1$ (-1 [EV], Cold white) & 0.073 & 0.031 & 0.016    \\
      $\hat{\myvector{R}}_2$ (-1 [EV], White)      & 0.072 & 0.029 & 0.016    \\
      $\hat{\myvector{R}}_3$ (-1 [EV], Warm white) & 0.076 & 0.030 & 0.016    \\
      $\hat{\myvector{R}}_4$ (0 [EV], Cold white)  & 0.029 & 0.008 & 0.004    \\
      $\hat{\myvector{R}}_5$ (0 [EV], White)       & 0.000 & 0.000 & 0.000    \\
      $\hat{\myvector{R}}_6$ (0 [EV], Warm white)  & 0.028 & 0.007 & 0.004    \\
      $\hat{\myvector{R}}_7$ (1 [EV], Cold white)  & 0.098 & 0.057 & 0.024    \\
      $\hat{\myvector{R}}_8$ (1 [EV], White)       & 0.099 & 0.061 & 0.023    \\
      $\hat{\myvector{R}}_9$ (1 [EV], Warm white)  & 0.101 & 0.068 & 0.026    \\ \hline\hline
    \end{tabular}
    }
    \label{tab:reflect_dssim}
  \end{table}

  Tables \ref{tab:reconst}, \ref{tab:reconst_mse}, and \ref{tab:reconst_dssim}
  show the PSNR, MSE, and DSSIM values between original image $\myvector{I}_i$ and reconstructed image $\hat{\myvector{I}}_i$
  for each $i$ for the three methods, where these values were averaged over the 45 image sets
  that we used in this simulation.
  From Table \ref{tab:reconst},
  we confirmed that the proposed network outperformed USI3D in terms of the reconstruction consistency
  although Bell's method provided the highest PSNR.
  This result indicates that current DNN-based methods such as USI3D
  have a limited performance in terms of reconstruction consistency
  compared with traditional optimization-based methods such as Bell's method.
  The proposed network partially overcomes this limited performance of DNN-based methods.
  This trend was also confirmed from Tables \ref{tab:reconst_mse} and \ref{tab:reconst_dssim}.
  \begin{table}[t]
    \centering
    \caption{PSNR values between $\myvector{I}_i$ and $\hat{\myvector{I}}_i$
      averaged over 45 image sets.
      Top: PSNR between $\myvector{I}_1$ and $\hat{\myvector{I}_1}$,
      Bottom: PSNR between $\myvector{I}_9$ and $\hat{\myvector{I}_9}$}
    {\footnotesize
    \begin{tabular}{l|ccc} \hline\hline
                  & Bell~\cite{bell2014intrinsic} & USI3D~\cite{liu2020unsupervised} & Proposed\\ \hline
      $\hat{\myvector{I}}_1$ (-1 {[}EV{]}, Cold white) & 46.53 & 20.43 & 21.75 \\
      $\hat{\myvector{I}}_2$ (-1 {[}EV{]}, White)      & 46.47 & 20.64 & 21.81 \\
      $\hat{\myvector{I}}_3$ (-1 {[}EV{]}, Warm white) & 46.55 & 20.66 & 21.78 \\
      $\hat{\myvector{I}}_4$ (0 {[}EV{]}, Cold white)  & 46.50 & 16.82 & 20.27 \\
      $\hat{\myvector{I}}_5$ (0 {[}EV{]}, White)       & 46.25 & 17.08 & 20.36 \\
      $\hat{\myvector{I}}_6$ (0 {[}EV{]}, Warm White)  & 46.19 & 17.14 & 20.37 \\
      $\hat{\myvector{I}}_7$ (+1 {[}EV{]}, Cold white) & 45.47 & 14.72 & 20.34 \\
      $\hat{\myvector{I}}_8$ (+1 {[}EV{]}, White)      & 45.38 & 14.76 & 20.40 \\
      $\hat{\myvector{I}}_9$ (+1 {[}EV{]}, Warm white) & 45.47 & 14.64 & 20.41 \\ \hline\hline
    \end{tabular}
    }
    \label{tab:reconst}
  \end{table}
  \begin{table}[t]
    \centering
    \caption{MSE values between $\myvector{I}_i$ and $\hat{\myvector{I}}_i$
      averaged over 45 image sets.
      Top: MSE between $\myvector{I}_1$ and $\hat{\myvector{I}_1}$,
      Bottom: MSE between $\myvector{I}_9$ and $\hat{\myvector{I}_9}$}
    {\footnotesize
    \begin{tabular}{l|ccc} \hline\hline
                  & Bell~\cite{bell2014intrinsic} & USI3D~\cite{liu2020unsupervised} & Proposed \\ \hline
      $\hat{\myvector{I}}_1$ (-1 [EV], Cold white) & 0.000 & 0.014 & 0.008    \\
      $\hat{\myvector{I}}_2$ (-1 [EV], White)      & 0.000 & 0.014 & 0.008    \\
      $\hat{\myvector{I}}_3$ (-1 [EV], Warm white) & 0.000 & 0.014 & 0.008    \\
      $\hat{\myvector{I}}_4$ (0 [EV], Cold white)  & 0.000 & 0.023 & 0.009    \\
      $\hat{\myvector{I}}_5$ (0 [EV], White)       & 0.000 & 0.022 & 0.009    \\
      $\hat{\myvector{I}}_6$ (0 [EV], Warm white)  & 0.000 & 0.022 & 0.009    \\
      $\hat{\myvector{I}}_7$ (1 [EV], Cold white)  & 0.000 & 0.036 & 0.008    \\
      $\hat{\myvector{I}}_8$ (1 [EV], White)       & 0.000 & 0.036 & 0.008    \\
      $\hat{\myvector{I}}_9$ (1 [EV], Warm white)  & 0.000 & 0.037 & 0.008    \\ \hline\hline
    \end{tabular}
    }
    \label{tab:reconst_mse}
  \end{table}
  \begin{table}[t]
    \centering
    \caption{DSSIM values between $\myvector{I}_i$ and $\hat{\myvector{I}}_i$
      averaged over 45 image sets.
      Top: DSSIM between $\myvector{I}_1$ and $\hat{\myvector{I}_1}$,
      Bottom: DSSIM between $\myvector{I}_9$ and $\hat{\myvector{I}_9}$}
    {\footnotesize
    \begin{tabular}{l|ccc} \hline\hline
                  & Bell~\cite{bell2014intrinsic} & USI3D~\cite{liu2020unsupervised} & Proposed \\ \hline
      $\hat{\myvector{I}}_1$ (-1 [EV], Cold white) & 0.001 & 0.161 & 0.121    \\
      $\hat{\myvector{I}}_2$ (-1 [EV], White)      & 0.001 & 0.162 & 0.122    \\
      $\hat{\myvector{I}}_3$ (-1 [EV], Warm white) & 0.001 & 0.165 & 0.124    \\
      $\hat{\myvector{I}}_4$ (0 [EV], Cold white)  & 0.001 & 0.204 & 0.101    \\
      $\hat{\myvector{I}}_5$ (0 [EV], White)       & 0.001 & 0.204 & 0.102    \\
      $\hat{\myvector{I}}_6$ (0 [EV], Warm white)  & 0.001 & 0.206 & 0.103    \\
      $\hat{\myvector{I}}_7$ (1 [EV], Cold white)  & 0.001 & 0.239 & 0.085    \\
      $\hat{\myvector{I}}_8$ (1 [EV], White)       & 0.001 & 0.240 & 0.086    \\
      $\hat{\myvector{I}}_9$ (1 [EV], Warm white)  & 0.001 & 0.242 & 0.087    \\ \hline\hline
    \end{tabular}
    }
    \label{tab:reconst_dssim}
  \end{table}

\subsection{Simulation using MIT intrinsic images dataset}
\subsubsection{Experimental conditions}
\begin{figure}[t]
      \centering
  	  \begin{subfigure}[t]{0.3\hsize}
        \centering
  		  \includegraphics[width=\columnwidth]{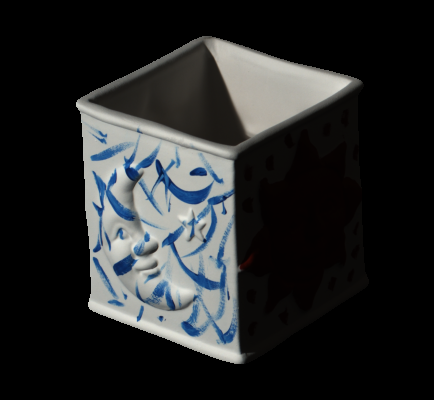}
        \caption{Original image}
      \end{subfigure}
      \begin{subfigure}[t]{0.3\hsize}
        \centering
  		  \includegraphics[width=\columnwidth]{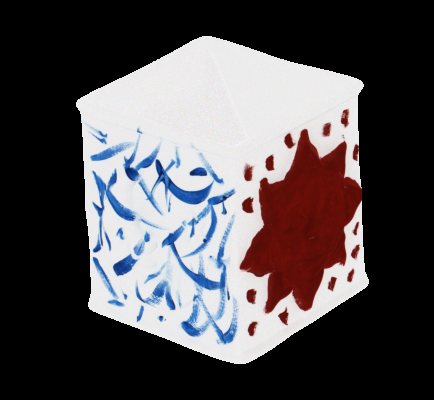}
        \caption{Reflectance}
      \end{subfigure}
  	  \begin{subfigure}[t]{0.3\hsize}
        \centering
  		  \includegraphics[width=\columnwidth]{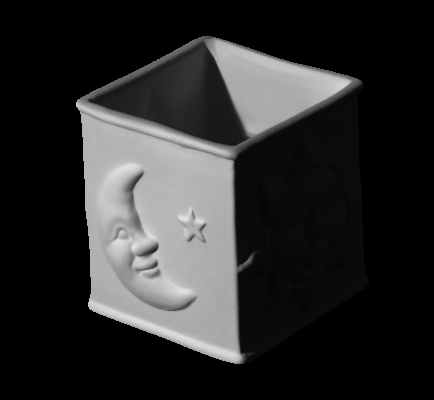}
        \caption{Shading}
      \end{subfigure}\\
      \caption{Groundtruth image ``box'' in Mit intrinsic image dataset~\cite{grosse2009groundtruth}
        \label{fig:mit}}
  \end{figure}

  To further study features of the proposed network,
  we conducted simulation experiments
  using the MIT intrinsic images dataset~\cite{grosse2009groundtruth},
  which is a typical dataset for intrinsic image decomposition.
  An example of an image ``box'' in the dataset is shown in Fig. \ref{fig:mit}.
  The MIT intrinsic images dataset provides not only original images
  but also the corresponding groundtruth reflectance and shading
  for each image.
  For constructing the dataset,
  computer graphics techniques were utilized.
  
  Because groundtruth reflectance and shading components
  are available for each image,
  in this experiment,
  we used the local mean squared error (LMSE) score~\cite{grosse2009groundtruth}
  to evaluate the performance of the proposed network.
  Here, a smaller LMSE score value means higher similarity
  between an estimation and the corresponding groundtruth.

  Other conditions were the same as those in Section \ref{sec:simulation}.\ref{subsec:raw_simulation}

\subsubsection{Expeirmental results}
\begin{figure}[t]
      \centering
  	  \begin{subfigure}[t]{0.3\hsize}
        \centering
  		  \includegraphics[width=\columnwidth]{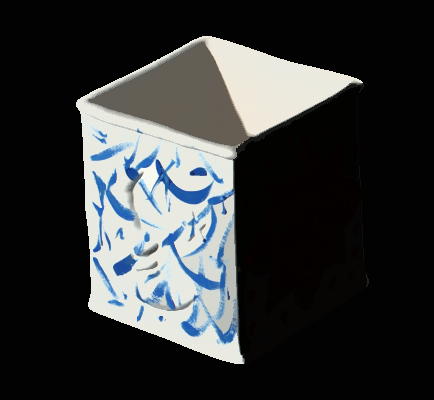}
        \caption{Bell~\cite{bell2014intrinsic}}
      \end{subfigure}
      \begin{subfigure}[t]{0.3\hsize}
        \centering
  		  \includegraphics[width=\columnwidth]{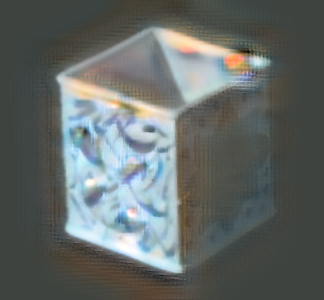}
        \caption{USI3D~\cite{liu2020unsupervised}}
      \end{subfigure}
  	  \begin{subfigure}[t]{0.3\hsize}
        \centering
  		  \includegraphics[width=\columnwidth]{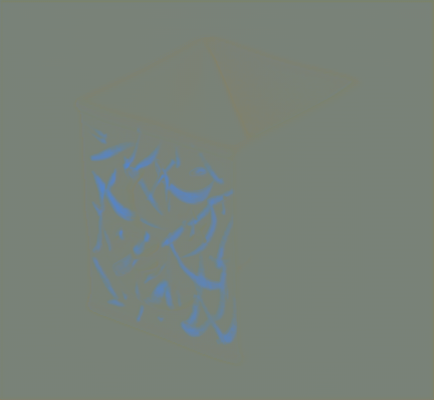}
        \caption{Proposed}
      \end{subfigure}\\
      \caption{Estimated reflectance for image ``box'' in Mit intrinsic image dataset
        \label{fig:mit_est_reflect}}
  \end{figure}

\begin{figure}[t]
      \centering
  	  \begin{subfigure}[t]{0.3\hsize}
        \centering
  		  \includegraphics[width=\columnwidth]{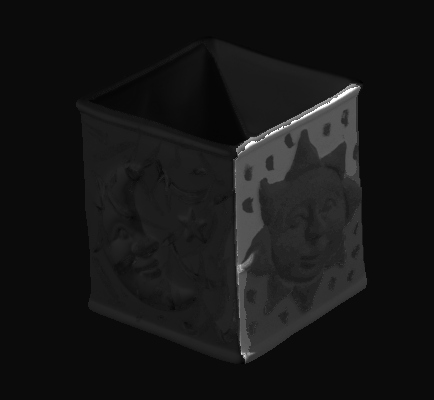}
        \caption{Bell~\cite{bell2014intrinsic}}
      \end{subfigure}
      \begin{subfigure}[t]{0.3\hsize}
        \centering
  		  \includegraphics[width=\columnwidth]{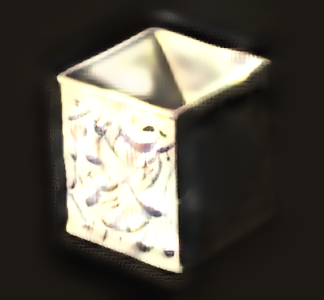}
        \caption{USI3D~\cite{liu2020unsupervised}}
      \end{subfigure}
  	  \begin{subfigure}[t]{0.3\hsize}
        \centering
  		  \includegraphics[width=\columnwidth]{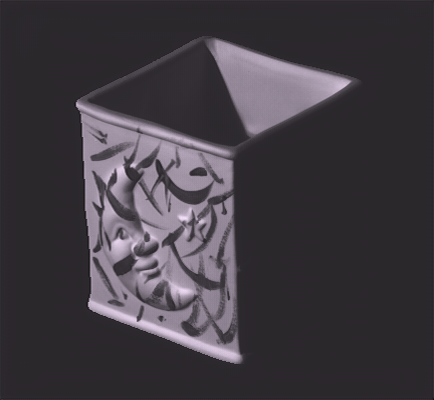}
        \caption{Proposed}
      \end{subfigure}\\
      \caption{Estimated shading for image ``box'' in Mit intrinsic image dataset
        \label{fig:mit_est_shading}}
  \end{figure}

  Figures \ref{fig:mit_est_reflect} and \ref{fig:mit_est_shading}
  show estimated reflectance and shading components, respectively,
  for an image ``box'' in MIT intrinsic images dataset
  estimated by Bell's method, USI3D, and the proposed network.
  As shown in Fig. \ref{fig:mit_est_reflect},
  the reflectance component estimated by the proposed network
  was significantly different from the groundtruth one shown in Fig. \ref{fig:mit}.
  This is because the proposed network was trained
  with luminance normalization of the reflectance
  in Eq. (\ref{eq:r_loss}).
  When there are no constraints on the luminance of reflectance,
  intrinsic image decomposition suffers
  from scale indeterminacy.
  In this case, when an image $\myvector{I}$ is decomposed
  as $\myvector{I}(x, y) = \myvector{S}(x, y) \odot \myvector{R}(x, y)$
  there is another solution $\myvector{I}(x, y) = (a \myvector{S}(x, y)) \odot (a^{-1} \myvector{R}(x, y))$,
  where $a > 0$.
  This scale indeterminacy may be solved
  by supervised learning using a large high-quality dataset.
  However, it is difficult to prepare the ground truth of reflectance and shading
  in real scenes.
  For this reason,
  we aim at intrinsic image decomposition by unsupervised learning,
  and the proposed network was trained with the luminance normalization
  to remove the scale indeterminacy.
  By using the normalization,
  the proposed network estimated shading components
  more accurately than the conventional methods,
  as shown in Fig. \ref{fig:mit_est_shading}.
  This is also confirmed by the fact that
  the proposed network provided LMSE scores
  as well as the conventional methods, as in Table \ref{tab:lmse}.
  \begin{table}[t]
    \centering
    \caption{LMSE scores for 16 images in MIT intrinsic dataset\label{tab:lmse}}
    {\footnotesize
    \begin{tabular}{l|ccc} \hline\hline
                  & Bell~\cite{bell2014intrinsic} & USI3D~\cite{liu2020unsupervised} & Proposed \\ \hline
      Box      & 0.210 & 0.056 & 0.067 \\
      Cup1     & 0.088 & 0.045 & 0.052 \\
      Cup2     & 0.130 & 0.062 & 0.085 \\
      Deer     & 0.173 & 0.091 & 0.109 \\
      Dinosaur & 0.133 & 0.072 & 0.095 \\
      Frog1    & 0.215 & 0.053 & 0.075 \\
      Frog2    & 0.269 & 0.114 & 0.059 \\
      Panther  & 0.098 & 0.045 & 0.062 \\
      Paper1   & 0.051 & 0.026 & 0.025 \\
      Paper2   & 0.048 & 0.030 & 0.032 \\
      Raccoon  & 0.143 & 0.049 & 0.040 \\
      Squirrel & 0.208 & 0.061 & 0.071 \\
      Sun      & 0.064 & 0.042 & 0.052 \\
      Teabag1  & 0.098 & 0.179 & 0.197 \\
      Teabag2  & 0.092 & 0.129 & 0.161 \\
      Turtle   & 0.166 & 0.078 & 0.052 \\ \hdashline
      Mean (16 images)    & 0.137 & 0.071 & 0.077 \\ \hline\hline
    \end{tabular}
    }
  \end{table}

\section{Conclusion}
  In this paper,
  we proposed a novel intrinsic image decomposition network
  considering reflectance consistency.
  In the proposed network,
  reconstruction consistency, reflectance consistency (brightness),
  and reflectance consistency (color) are considered
  by using a color-illuminant model
  and training the network with losses calculated
  from images taken under various illumination conditions.
  In addition, the proposed network can be trained in a self-supervised manner
  because various illumination conditions can easily be simulated.
  Experimental results show that our network can decompose images
  into reflectance and shading components while maintaining the reflectance consistencies
  in terms of both illumination-brightness and -colors.

  Since the proposed network can produce robust reflectance against changes of illumination conditions,
  the proposed network will contribute to varous color image processing such as
  image enhancement while preserving object color
  and more realistic white balance adjustment.

\section*{Financial Support}
  ``None.''

\section*{Statement of interest}
  ``None.''

\vskip2pc



\vskip2pc

\noindent \large \textbf{Biographies}

\vskip2pc

\noindent\normalsize\textbf{Yuma Kinoshita}
  received his B.Eng., M.Eng., and the Ph.D. degrees
  from Tokyo Metropolitan University, Japan,
  in 2016, 2018, and 2020 respectively.
  Since April 2020,
  he has been a project assistant professor
  at Tokyo Metropolitan University.
  His research interests are in the area
  of signal processing, image processing,
  and machine learning.
  He is a Member of IEEE, APSIPA, IEICE, and ASJ.
  He received the IEEE ISPACS Best Paper Award, in 2016,
  the IEEE Signal Processing Society Japan Student
  Conference Paper Award, in 2018,
  the IEEE Signal Processing Society Tokyo Joint Chapter
  Student Award, in 2018,
  the IEEE GCCE Excellent Paper Award (Gold Prize), in 2019,
  and the IWAIT Best Paper Award, in 2020.
  He was a Registration Chair of DCASE2020 Workshop.

\vskip2pc

\noindent\textbf{Hitoshi Kiya}
  received his B.E. and M.E. degrees from
  the Nagaoka University of Technology, Japan,
  in 1980 and 1982, respectively,
  and his Dr.Eng. degree from
  Tokyo Metropolitan University in 1987.
  In 1982, he joined Tokyo Metropolitan University,
  where he became a Full Professor in 2000.
  From 1995 to 1996, he attended The University of Sydney,
  Australia, as a Visiting Fellow.
  He is a fellow of IEEE, IEICE, and ITE.
  He was a recipient of numerous awards,
  including 10 best paper awards.
  He served as the President of APSIPA from 2019 to 2020,
  and the Regional Director-at-Large for Region 10 of
  the IEEE Signal Processing Society from 2016 to 2017.
  He was also the President of the IEICE Engineering
  Sciences Society from 2011 to 2012.
  He has been an editorial board member of eight journals,
  including IEEE Transactions on Signal Processing,
  IEEE Transactions on Image Processing,
  and IEEE Transactions on Information Forensics
  and Security, and a member of nine technical committees,
  including the APSIPA Image, Video,
  and Multimedia Technical Committee (TC),
  and IEEE Information Forensics and Security TC.
  He has organized a lot of international conferences
  in such roles as the TPC Chair of IEEE ICASSP 2012
  and as the General Co-Chair of IEEE ISCAS 2019.
\end{document}